\documentclass[]{thinking_with_video}

\usepackage{helvet}
\usepackage[utf8]{inputenc}
\usepackage[T1]{fontenc}
\usepackage{lmodern}
\usepackage{amsmath,amssymb,amsfonts}
\usepackage{natbib}
\usepackage{graphicx}
\usepackage{subcaption}
\usepackage{caption}
\usepackage{float}
\usepackage{booktabs}
\usepackage{multirow}
\usepackage{makecell}
\usepackage{array}
\usepackage{tabularx}
\usepackage[table]{xcolor}
\usepackage{enumitem}
\usepackage{pifont}
\usepackage[most]{tcolorbox}
\usepackage[toc,page,header]{appendix}
\usepackage{titletoc}
\usepackage{minitoc}
\usepackage{microtype}
\usepackage{listings}
\usepackage{hyperref}
\usepackage{url}
\hypersetup{
  colorlinks=true,
  linkcolor=seedblue,
  citecolor=seedblue,
  urlcolor=seedblue
}

\definecolor{priorbenchblue}{RGB}{246,248,253}
\definecolor{oursblue}{RGB}{224,236,255}

\definecolor{takeawayblue}{RGB}{48,105,255}
\newtcolorbox{takeawaybox}{
  enhanced,
  colback=takeawayblue!4,
  colframe=takeawayblue,
  boxrule=0.8pt,
  arc=2mm,
  left=2.5mm,
  right=2.5mm,
  top=1.5mm,
  bottom=1.5mm,
  before skip=5pt,
  after skip=7pt,
  fontupper=\small
}

\newtcolorbox{promptbox}[2][]{
  enhanced,
  breakable,
  colback=white,
  coltext=black,
  colframe=black!60,
  colbacktitle=black,
  coltitle=white,
  fonttitle=\bfseries,
  title={#2},
  arc=2mm,
  boxrule=0.5pt,
  top=6pt,
  bottom=6pt,
  left=8pt,
  right=8pt,
  before upper={\setlength{\parindent}{0pt}\setlength{\parskip}{0.7ex}},
  #1
}

\lstdefinestyle{promptstyle}{
  basicstyle=\ttfamily\scriptsize,
  breaklines=true,
  breakatwhitespace=false,
  columns=fullflexible,
  keepspaces=true,
  showstringspaces=false,
  frame=none,
  backgroundcolor=\color{white},
  xleftmargin=0pt,
  xrightmargin=0pt,
  aboveskip=0pt,
  belowskip=0pt,
  tabsize=2
}

\DeclareRobustCommand{\greencheck}{\textcolor[RGB]{41,155,118}{\ding{51}}}
\DeclareRobustCommand{\redcross}{\textcolor[RGB]{190,45,45}{\ding{55}}}

\newcommand{\titlelogo}{%
\raisebox{-0.25em}{\includegraphics[height=2.25em]{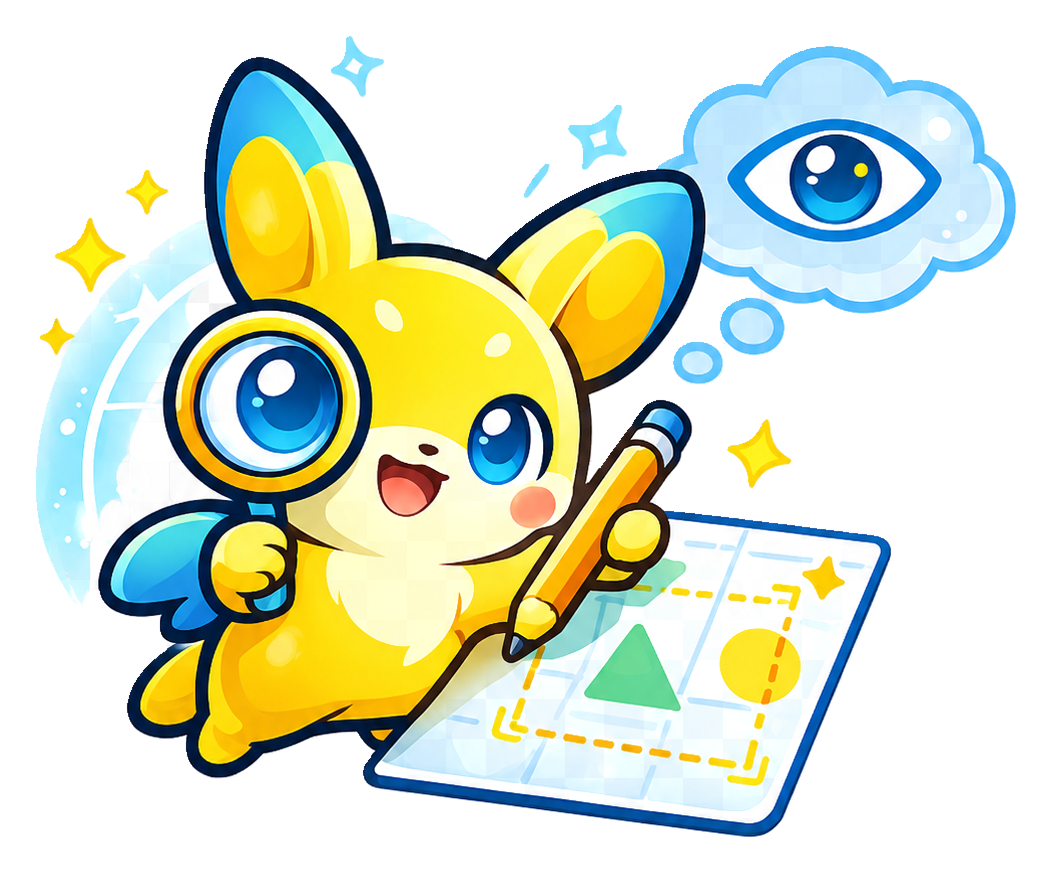}}%
}

\title{\titlelogo\hspace{0.1em}See2Think: Do Multimodal Models Really Use Intermediate Visual States?}

\renewcommand{\authorfont}{\fontsize{10}{10.5}\selectfont}
\renewcommand{\affiliationfont}{\fontsize{9.2}{11.2}\selectfont}

\author{
\mbox{Siyu Yan\textsuperscript{1,3,$\dagger$}},
\mbox{Zhuoran Yan\textsuperscript{2,$\dagger$}},
\mbox{Haiying Xu\textsuperscript{3,4,$\dagger$}},
\mbox{Panhao Zhou\textsuperscript{2}},
\mbox{Jingyu Chen\textsuperscript{2}},
\mbox{Chenhao Ji\textsuperscript{3}},
\mbox{Shuo Cao\textsuperscript{3,5}},
\mbox{Yongheng Zhang\textsuperscript{2}},
\mbox{Haoze Liu\textsuperscript{3}},
\mbox{Siyu Zhang\textsuperscript{3,6}},
\mbox{Xiwen Gu\textsuperscript{7}},
\mbox{Yihao Liu\textsuperscript{3}},
\mbox{Alex Jinpeng Wang\textsuperscript{2,$\S$}}
}

\renewcommand{\affiliationlist}{%
{\affiliationfont
$^{1}$The Hong Kong University of Science and Technology
\qquad
$^{2}$Central South University\\[0.6mm]
$^{3}$Shanghai AI Laboratory
\qquad
$^{4}$The Hong Kong University of Science and Technology (Guangzhou)\\[0.6mm]
$^{5}$University of Science and Technology of China
\qquad
$^{6}$Fudan University
\qquad
$^{7}$Wuhan University
}%
}

\abstract{
Multimodal large language models increasingly use sketches, annotations, tools,
and intermediate images during reasoning, but it remains unclear whether they
truly rely on these visual states. Existing benchmarks are limited both by task
collections with narrow coverage or partially text-solvable samples and by
evaluations that emphasize final answers without diagnosing how intermediate
visual states are generated, rendered, and used.
We introduce \textbf{See2Think}, a unified evaluation framework comprising
\textbf{See2ThinkBench} and \textbf{Visual Action-of-Thought (VAoT)}.
See2ThinkBench contains \textbf{1,200 open-ended, visually dependent problems}
across 12 task categories spanning 2D structured, 3D scene, and real-world
reasoning. VAoT records textual thoughts, visual actions, rendered states, and
subsequent reasoning under four controlled inference settings.
Evaluating representative proprietary and open-source multimodal models, we
find that visual reasoning is strongly model- and environment-dependent, with
no single setting consistently dominating across tasks. Process analysis
further shows that models usually select relevant visual operations, while
faithful rendering remains the clearest bottleneck and high feedback uptake
does not necessarily translate into accuracy gains. Under task-relevant
corrupted feedback, models exhibit behavioral dependence on visual states,
with accuracy dropping by \textbf{over 10 percentage points} in 3D scenes.
See2Think disentangles visual-state utility from behavioral dependence and
provides a unified framework for evaluating whether multimodal models genuinely
use intermediate visual states.
}

\checkdata[Website]{\url{https://sgysy.github.io/seetothink/}}
\checkdata[Repository]{\url{https://github.com/CSU-JPG/See2Think}}

\begin{document}
\maketitle
\renewcommand{\thefootnote}{}
\footnotetext{\textsuperscript{$\dagger$} Equal contribution.}
\footnotetext{\textsuperscript{$\S$} Corresponding author.}
\renewcommand{\thefootnote}{\arabic{footnote}}

\vspace{-1.5em}

\section{Introduction}
\label{sec:introduction}

\begin{figure*}[t]
\centering
\includegraphics[width=\textwidth]{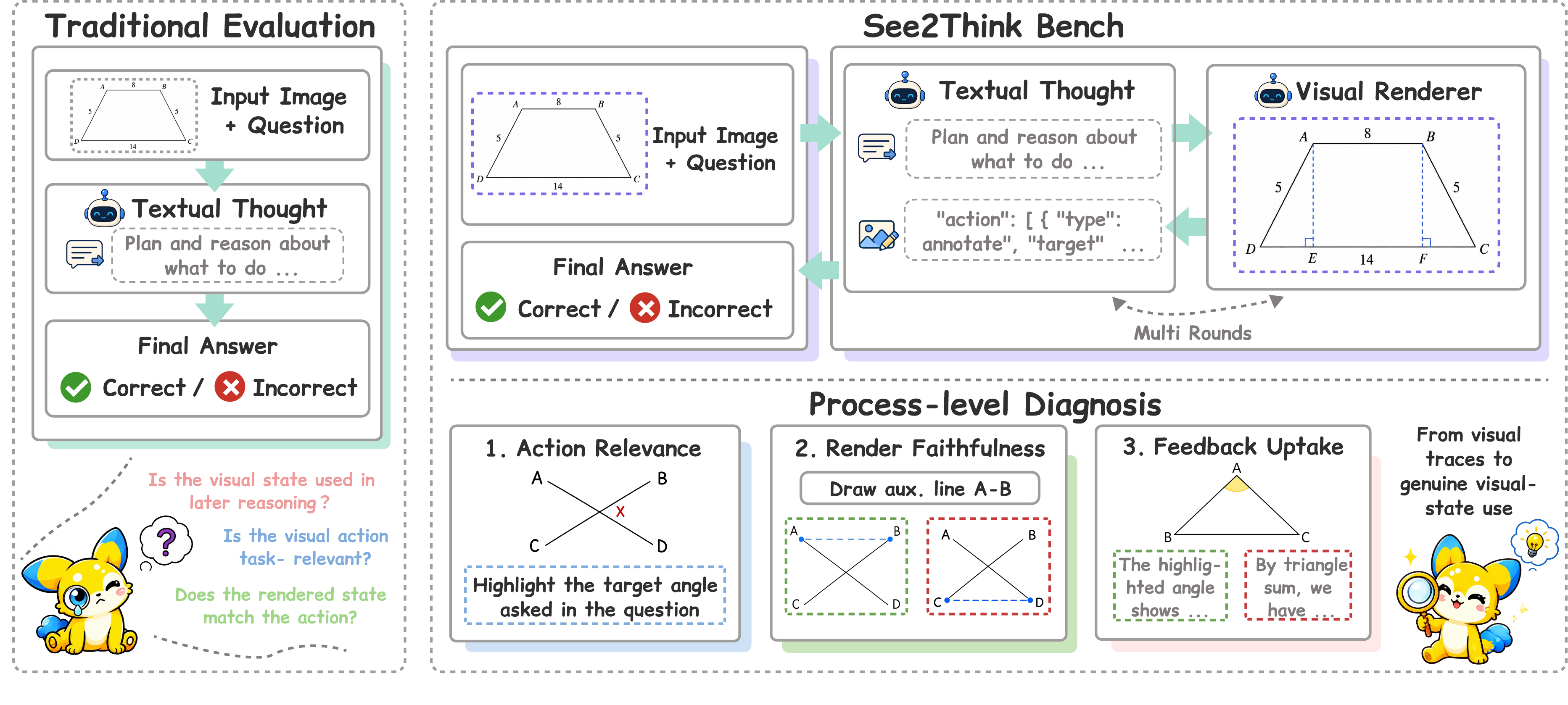}
\caption{\textbf{See2Think versus prior visual-reasoning benchmarks.}
Prior benchmarks mainly evaluate final answers or visual-trace quality.
See2Think diagnoses action relevance, render faithfulness, and feedback uptake.
Controlled interventions test whether models rely on the returned visual state.}
\label{fig:see2think_main}
\end{figure*}

Chain-of-Thought (CoT) reasoning improves language-model reasoning by
externalizing intermediate textual steps~\citep{wei2022chain}. Many multimodal
problems, however, are not naturally solved through text alone. Humans draw
auxiliary lines, mark relevant regions, compare structures, and sketch
intermediate states to externalize spatial and perceptual relations. Recent
multimodal systems similarly generate visual scratchpads, invoke visual tools,
highlight or crop image regions, and construct intermediate images during
reasoning~\citep{zhang2023multimodalcot,hu2024visualsketchpad,zheng2025deepeyes,
su2025openthinkimg,shi2025mathcanvas,qiao2025vthinker}.

Building on this idea, recent work has rapidly expanded both the methods and
evaluations for thinking with images. Models now crop, annotate, sketch,
manipulate, or generate intermediate visual states through external tools,
interactive image generation, continuous visual actions, and repeated visual
access
\citep{hu2024visualsketchpad,zheng2025deepeyes,
su2025openthinkimg,shi2025mathcanvas,qiao2025vthinker,
li2026reliablethinking,zhao2026nvcot,yang2026walkthetalk,
li2026s1vl,hu2026tvicot}.
Benchmarks such as MIRA, ViC-Bench, TWI-PRMBench, and TwiFF-Bench evaluate
oracle visual clues, free-form interleaved states, process quality, or dynamic
future-frame reasoning
\citep{zhou2025mira,wu2025vicbench,
zhou2026thinkwithimages,liu2026twiff}.
Recent diagnostic studies further question whether tool calls or generated
thinking images are actually answer-critical
\citep{guo2026toolbenefit,yang2026howwhatimagine}.

Two limitations nevertheless remain. At the data level, existing task
collections may have limited visual diversity and are not always screened for
text-only shortcuts. At the evaluation level, standard multimodal benchmarks
emphasize final answers~\citep{yue2024mmmu,lu2023mathvista}, while multimodal
CoT evaluations mainly assess textual rationales~\citep{jiang2025mmecot}.
Visual-thinking benchmarks introduce sketches, tools, or intermediate images,
but are commonly evaluated through end-task performance, oracle-provided
visual clues, or aggregate process scores without matched interventions.
Consequently, existing evaluations do not jointly determine whether a
self-generated visual action is task-relevant, whether the returned state
faithfully realizes that action, and whether subsequent reasoning behaviorally
depends on the returned visual evidence.

We introduce \textbf{See2Think}, a unified framework for evaluating whether
multimodal models genuinely think with intermediate visual states. See2Think
contains two complementary components. \textbf{See2ThinkBench} provides
\textbf{1.2K visually dependent reasoning problems} across 12 task categories
and three visual environments: 2D structured reasoning, 3D scene reasoning,
and real-world visual reasoning. Caption-only solvability filtering reduces
text-dominant shortcuts. Most tasks use free-form answer generation, while a
small subset retains structured candidate sets inherited from source benchmarks
when those candidates form part of the reasoning problem.
\textbf{Visual Action-of-Thought (VAoT)} provides the inference-time protocol:
it records alternating textual thoughts, visual actions, rendered states, and
subsequent reasoning. Controlled variants compare CoT, action planning without
rendering, standard closed-loop VAoT, and task-relevant corrupted feedback
designed to probe behavioral dependence.
Together, these components support both outcome-level evaluation and diagnosis
of action selection, visual execution, and downstream state use.

Our contributions are:
\begin{enumerate}[leftmargin=*]
\item We formulate genuine visual-state use as a \textbf{controlled diagnostic
problem} and introduce \textbf{See2Think}, which distinguishes whether an
intermediate visual state is useful from whether model behavior actually
depends on it.

\item We construct \textbf{See2ThinkBench}, a benchmark of 1.2K visually
dependent reasoning problems across 12 task categories and three visual
environments. Caption-only solvability filtering reduces text-dominant
shortcuts, while answer-format normalization supports consistent evaluation
across free-form and structured outputs.

\item We introduce \textbf{VAoT}, an observable and intervenable protocol for
recording visual actions, rendered states, and their downstream use. Experiments
across four representative proprietary and open-source models reveal no
universally optimal visual-reasoning strategy, identify a persistent bottleneck
after action selection, and expose when rendered feedback helps, harms, or
merely decorates the reasoning process.
\end{enumerate}

\section{See2ThinkBench}
\label{sec:see2thinkbench}

\begin{figure*}[t]
\centering
\includegraphics[width=\textwidth]{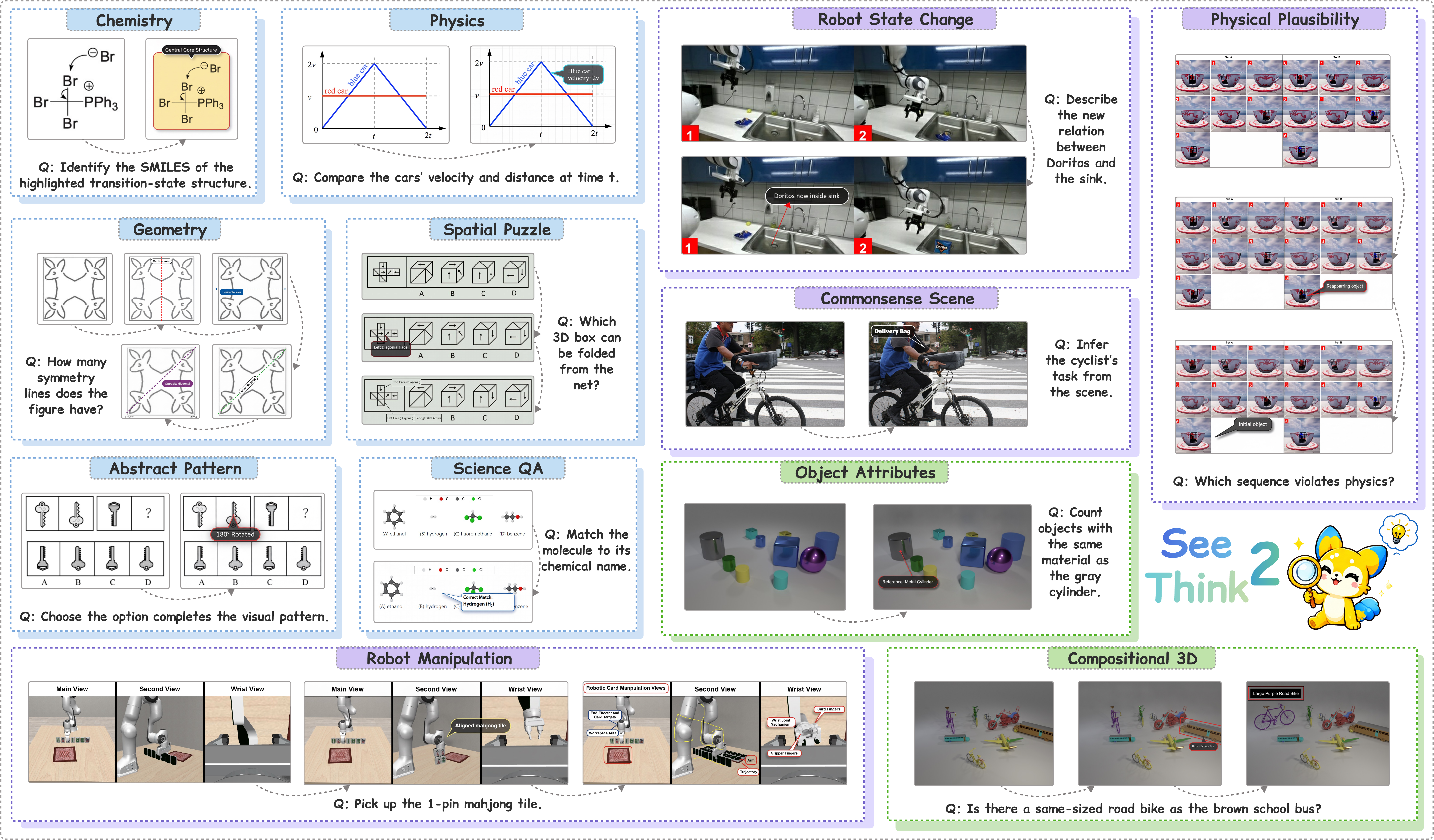}
\caption{\textbf{Overview of See2ThinkBench.}
Representative examples from 12 task categories spanning three visual reasoning
levels: 2D structured reasoning, 3D scene reasoning, and real-world visual
reasoning.}
\label{fig:see2thinkbench_overview}
\end{figure*}

As the data component of See2Think, See2ThinkBench supports controlled
outcome- and process-level diagnosis of intermediate visual states in
multimodal reasoning. Its construction follows
three principles: each problem should require visual evidence, naturally
support meaningful visual operations, and permit matched outcome- and
process-level evaluation.

\subsection{Task Design and Data Sources}
\label{sec:task_design}

See2ThinkBench contains \textbf{1.2K visually dependent reasoning samples}
from \textbf{12 task categories}, with 100 samples in each category. As shown in
Figure~\ref{fig:see2thinkbench_overview}, these categories are organized into
three visual reasoning levels: \textit{2D structured reasoning},
\textit{3D scene reasoning}, and \textit{real-world visual reasoning}.

\textbf{2D structured reasoning.}
This level comprises six task categories: \textit{Geometry},
\textit{Spatial Puzzle}, \textit{Physics}, \textit{Chemistry},
\textit{Science QA}, and \textit{Abstract Pattern}. These problems expose
explicit structures such as geometric relations, symbolic diagrams, temporal
curves, molecular configurations, and visual transformations. Useful visual
operations include drawing auxiliary lines, marking intersections, tracing
relations, and annotating symbolic evidence.

\textbf{3D scene reasoning.}
This level comprises \textit{Object Attributes} and
\textit{Compositional 3D}. The model must identify target objects and reason
about attributes, relative positions, counting, comparison, and multi-hop
spatial relations. Useful operations include marking objects, isolating
regions, tracking relation chains, and filtering distractors.

\textbf{Real-world visual reasoning.}
This level comprises \textit{Robot Manipulation}, \textit{Robot State Change},
\textit{Visual Commonsense}, and \textit{Intuitive Physics}. These tasks
require semantic grounding, object-state tracking, action understanding,
physical plausibility, affordances, and scene-level commonsense. Useful visual
operations are less focused on exact geometric construction and more on
isolating perceptual evidence and tracking changes across observations.

The source datasets used to instantiate these categories are documented in
Appendix~\ref{app:source_sampling}; the main paper reports results by semantic
task category rather than by source-dataset name.

Table~\ref{tab:benchmark_comparison} compares See2Think with representative
benchmarks for visual intermediate reasoning. MIRA evaluates the benefit of
annotated intermediate visualizations, ViC-Bench supports free-form
visual-interleaved trajectories, TWI-PRMBench focuses on process reward
modeling, and TwiFF-Bench extends visual reasoning to dynamically generated
future states. See2Think differs by combining explicit text-shortcut filtering
with self-generated closed-loop visual states, stage-wise diagnosis of action,
rendering, and feedback use, and task-relevant corrupted-feedback interventions.

\begin{table*}[t]
\centering
\scriptsize
\setlength{\tabcolsep}{5.5pt}
\renewcommand{\arraystretch}{1.18}

\caption{\textbf{Comparison with representative benchmarks and evaluation
frameworks for visual intermediate reasoning.}
TWI-PRMBench denotes ThinkWithImages-PRMBench.
\# Tasks denotes the number of reported task categories or subcategories.
Feedback Intervention indicates controlled modification of intermediate
visual feedback.
Action--Render--Use Diagnosis indicates explicit evaluation of action
selection, visual execution, and downstream visual-state use.}
\label{tab:benchmark_comparison}

\resizebox{\textwidth}{!}{%
\begin{tabular}{lccccccc}
\toprule

\multirow[c]{2}{*}{\textbf{Benchmark}}
& \multirow[c]{2}{*}{\textbf{\# Samples}}
& \multirow[c]{2}{*}{\textbf{\# Tasks}}
& \textbf{Generated}
& \textbf{Closed-}
& \textbf{Process}
& \textbf{Feedback}
& \textbf{Action--Render--Use} \\

& &
& \textbf{IVS}
& \textbf{loop}
& \textbf{Eval.}
& \textbf{Interv.}
& \textbf{Diagnosis} \\

\midrule

\rowcolor{priorbenchblue}
MIRA~\citep{zhou2025mira}
& 546
& 20
& \redcross
& \redcross
& \redcross
& \redcross
& \redcross \\

\rowcolor{priorbenchblue}
ViC-Bench~\citep{wu2025vicbench}
& 2,751
& 4
& \greencheck
& \greencheck
& \greencheck
& \redcross
& \redcross \\

\rowcolor{priorbenchblue}
TWI-PRMBench~\citep{zhou2026thinkwithimages}
& 1,206
& 16
& \greencheck
& \redcross
& \greencheck
& \redcross
& \redcross \\

\rowcolor{priorbenchblue}
TwiFF-Bench~\citep{liu2026twiff}
& 1,078
& 3
& \greencheck
& \greencheck
& \greencheck
& \greencheck
& \redcross \\

\midrule

\rowcolor{oursblue}
\textbf{See2Think (Ours)}
& \textbf{1,200}
& \textbf{12}
& \greencheck
& \greencheck
& \greencheck
& \greencheck
& \greencheck \\

\bottomrule
\end{tabular}%
}

\end{table*}

\subsection{Data Construction}
\label{sec:data_construction}

\begin{figure*}[t]
\centering
\includegraphics[width=0.8\textwidth]{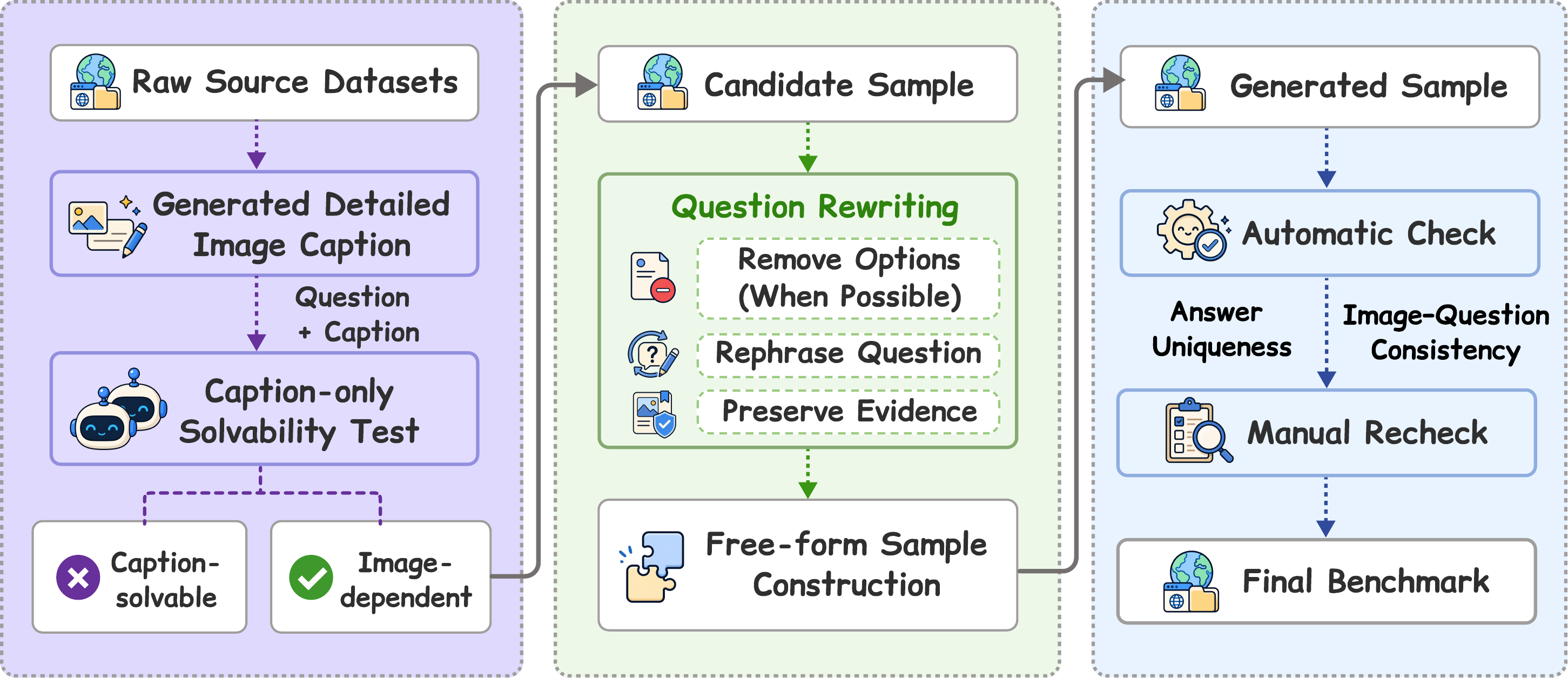}
\caption{\textbf{Construction pipeline.}
Candidate samples are filtered through caption-only answering, converted into
a unified answer-generation format, and subjected to automatic checks and
manual verification for visual dependency and answer quality.}
\label{fig:see2thinkbench_pipeline}
\end{figure*}

Figure~\ref{fig:see2thinkbench_pipeline} summarizes the construction pipeline.
Starting from candidate samples collected from existing visual reasoning
datasets, we filter out text-dominant cases, normalize the remaining tasks into
a unified answer-generation format, and apply automatic and manual quality
control for visual dependency, answer uniqueness, and image--question
consistency.

\textbf{Visual dependency filtering.}
A central concern in multimodal benchmark construction is
\textit{text-dominant bias}: some samples can be answered from the question,
language priors, or a sufficiently detailed textual description without
inspecting the original image. Such samples are unsuitable for diagnosing
intermediate visual-state use. For each candidate, we use a strong multimodal
model to generate a detailed caption conditioned on both the image and the
question. The captioner is instructed to describe only visually observable
evidence and not to answer the question or introduce unsupported inference.
We then provide a reasoning model with the question and caption while
withholding the original image, and query it five times. Samples answered
correctly in at least three trials are considered caption-solvable and removed;
the remaining samples are retained as more visually dependent candidates.
Prompting and model details are provided in the appendix.

\textbf{Answer-format conversion.}
Many source benchmarks contain multiple-choice or template-based questions,
which may introduce guessing shortcuts, answer priors, or option-elimination
behavior. Whenever possible, we remove textual answer-option lists and rewrite
the question as a free-form query while preserving the evidence required to
determine a unique answer. A small subset (127 of 1,200 samples) retains
structured candidate sets inherited from source tasks because the candidates
themselves form part of the reasoning problem; in some cases they appear in the
question text, and in others as panels embedded in the original image. These
instances are evaluated as structured-answer tasks, while the remaining 1,073
use free-form answers. For natural and embodied scenes, we additionally verify
that each question remains grounded in visually observable evidence.

\textbf{Quality control.}
The converted problems should require non-trivial reasoning beyond direct
visual lookup, admit a unique and reproducible answer, and remain consistent
with the associated image. We use a language model to generate candidate
rewrites, followed by automatic checks for answer uniqueness and
image--question consistency. Each retained sample is then manually rechecked
for visual dependency, ambiguity, and compatibility with visual-action
reasoning. Samples with underspecified answers, ambiguous visual references,
or insufficient visual grounding are revised or removed.

\subsection{Evaluation Targets}
\label{sec:evaluation_targets}

See2ThinkBench supports two complementary evaluation targets:
final-answer correctness and process-level quality. The concrete inference
conditions and intervention comparisons are introduced in
Section~\ref{sec:experiments}; here, we define the evaluation targets supported
by the benchmark.

\textbf{Final-answer correctness.}
Each sample provides a reference answer for outcome-level evaluation. For
structured-answer formats, we use exact matching after standard normalization.
For free-form predictions, we evaluate semantic equivalence between the model
prediction and the reference answer. This target supports conventional
accuracy-based comparison across models and inference conditions.

\textbf{Process-level quality.}
For complete VAoT trajectories, we additionally evaluate the quality of the
intermediate visual reasoning process. Because a trajectory may contain
multiple visual-action calls, we first identify the key visual step or steps
most relevant to the final reasoning process. These steps are then evaluated
along three dimensions: \textit{action relevance}, which measures whether the
requested visual operation targets task-relevant evidence; \textit{render
faithfulness}, which measures whether the rendered state correctly realizes
the requested operation; and \textit{feedback uptake}, which measures whether
subsequent reasoning uses information actually present in the rendered visual
state. For diagnostic analysis, incorrect trajectories can additionally be
assigned a primary failure source. Detailed scoring protocols are provided in Appendix~\ref{app:evaluation_protocols}, and human validation is reported in Section~\ref{sec:human_alignment}.
\section{Visual Action-of-Thought}
\label{sec:vaot_protocol}

\begin{figure*}[t]
\centering
\includegraphics[width=0.9\textwidth]{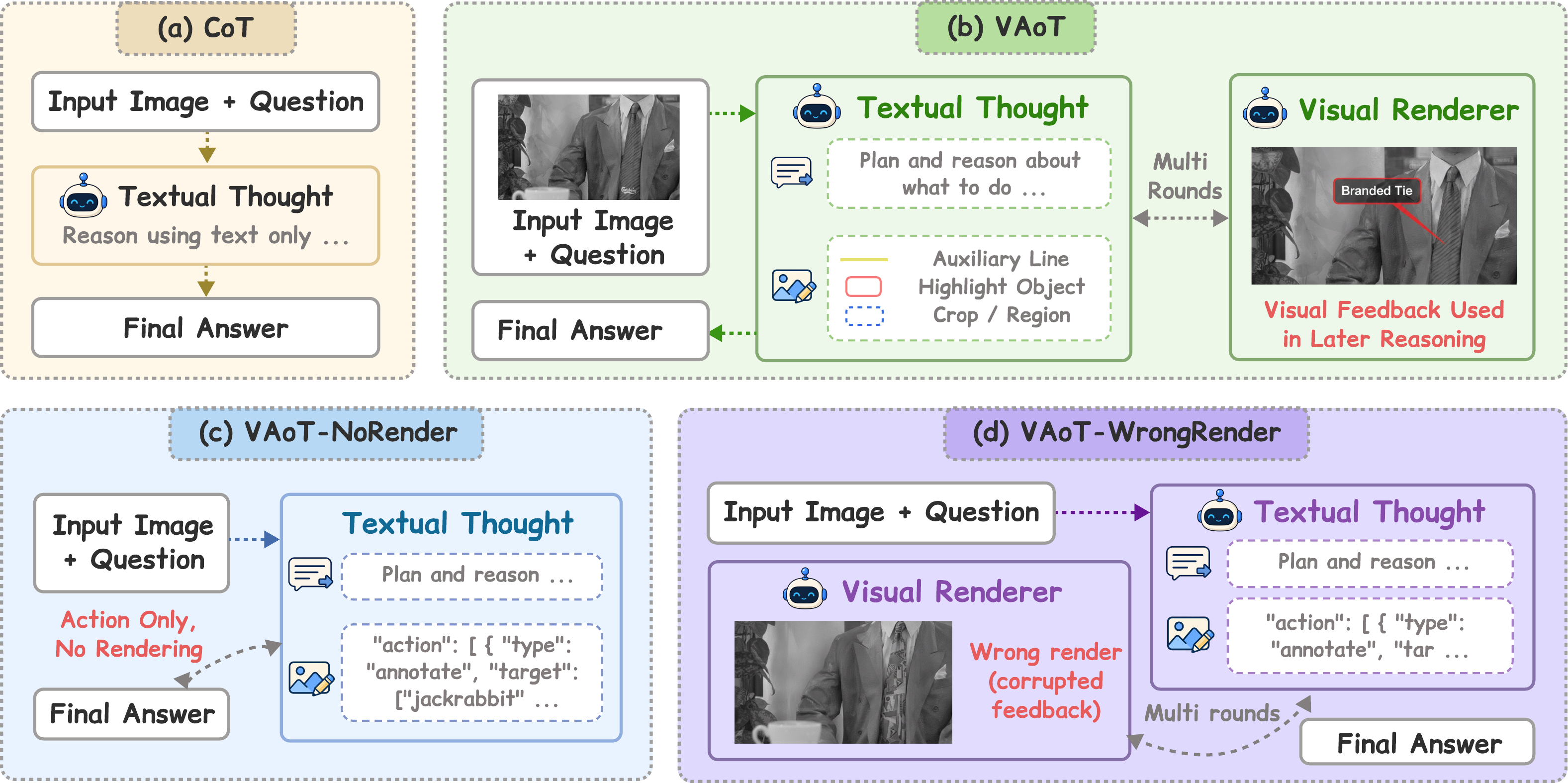}
\caption{\textbf{Visual Action-of-Thought.} VAoT interleaves textual thoughts,
visual actions, and rendered visual states. The model proposes a structured
action, the renderer applies a constrained edit to the current image, and the
resulting state is returned for subsequent reasoning.}
\label{fig:vaot_protocol}
\end{figure*}

As the protocol component of See2Think, Visual Action-of-Thought (VAoT)
records intermediate visual reasoning as an interleaved sequence of textual
thoughts, image-grounded actions, and rendered visual states. Because each
action can be performed and inspected externally, the trajectory supports
controlled intervention and process-level diagnosis. Figure~\ref{fig:vaot_protocol}
illustrates this closed-loop interaction and its controlled inference settings.

Given an input image $I$ and question $q$, the model produces a textual thought
$T_t$ and a structured visual action $A_t$ on the current visual state. The
action specifies an operation, normalized target coordinates, and optional
semantic content or style. An external renderer executes it and returns the
updated state $R_t$. The interaction repeats until a final answer is produced:
\begin{equation}
\mathcal{T}=\{(T_t,A_t,R_t)\}_{t=1}^{N}.
\end{equation}

Defining the accumulated history as
$\mathcal{H}_t=\{(T_i,A_i,R_i)\}_{i=1}^{t}$,
the next textual step is

\begin{equation}
T_{t\text{+}1}\sim M\left(I,q,\mathcal{H}_t\right).
\end{equation}

The action space includes highlighting objects, drawing auxiliary lines,
cropping or zooming into relevant regions, and annotating visual relations.
Each action must target visible image content or a visually grounded
construction, making the operation executable and externally inspectable.

\textbf{Renderer principle.}
The renderer is an external visual workspace that is constrained rather than a
task-solving agent. Given the current visual state and a structured action, it
executes the requested edit while preserving unrelated content. It receives
neither the reference answer nor a reasoning objective. The execution and
intervention conditions are defined in Section~\ref{sec:exp_setup}.

VAoT is an inference-time diagnostic protocol and requires no access to the internals of the model.

\section{Experiments}
\label{sec:experiments}

\begin{table*}[t]
\centering
\scriptsize
\setlength{\tabcolsep}{2.7pt}
\renewcommand{\arraystretch}{1.05}
\caption{\textbf{Overall accuracy on See2ThinkBench.}
All settings are evaluated on all 1,200 samples, with 100 samples from each task category.
The best result among CoT, VAoT-NoRender, and VAoT is highlighted in
bold; ties are highlighted jointly. VAoT-WrongRender is excluded from
best-score highlighting. Abbreviations: S-Puz. = Spatial Puzzle,
AbsPat = Abstract Pattern, ObjAttr = Object Attributes,
Comp3D = Compositional 3D, R-Man. = Robot Manipulation,
R-State = Robot State Change, V-Comm. = Visual Commonsense, and
IntPhys = Intuitive Physics.}
\label{tab:main_acc}
\resizebox{\textwidth}{!}{
\begin{tabular}{lcccccc|cc|cccc|c}
\toprule
\multirow{2}{*}{\textbf{Setting}}
& \multicolumn{6}{c|}{\textbf{2D Structured Reasoning}}
& \multicolumn{2}{c|}{\textbf{3D Scene Reasoning}}
& \multicolumn{4}{c|}{\textbf{Real-world Visual Reasoning}}
& \multirow{2}{*}{\textbf{Overall}} \\
\cmidrule(lr){2-7}\cmidrule(lr){8-9}\cmidrule(lr){10-13}
& \textbf{Geo.} & \textbf{S-Puz.} & \textbf{Phys.} & \textbf{Chem.} & \textbf{SciQA} & \textbf{AbsPat} & \textbf{ObjAttr} & \textbf{Comp3D} & \textbf{R-Man.} & \textbf{R-State} & \textbf{V-Comm.} & \textbf{IntPhys} & \\
\midrule
\multicolumn{14}{c}{\cellcolor{gray!12}\textit{\textbf{GPT-5.5}}} \\
CoT & \textbf{81.0} & \textbf{69.0} & \textbf{48.0} & 26.0 & 68.0 & 40.0 & 59.0 & \textbf{46.0} & 1.0 & \textbf{39.0} & 56.0 & 69.0 & \textbf{50.2} \\
VAoT-NoRender & 60.0 & 53.0 & \textbf{48.0} & 23.0 & 69.0 & \textbf{47.0} & \textbf{72.0} & 42.0 & \textbf{3.0} & 34.0 & 55.0 & \textbf{70.0} & 48.0 \\
VAoT & 46.0 & 41.0 & 46.0 & \textbf{28.0} & \textbf{72.0} & 39.0 & 68.0 & 37.0 & 1.0 & 38.0 & \textbf{61.0} & 58.0 & 44.6 \\
VAoT-WrongRender & 55.0 & 45.0 & 47.0 & 20.0 & 60.0 & 36.0 & 60.0 & 35.0 & 1.0 & 36.0 & 50.0 & 62.0 & 42.3 \\
\midrule
\multicolumn{14}{c}{\cellcolor{gray!12}\textit{\textbf{GPT-o3}}} \\
CoT & \textbf{35.0} & \textbf{32.0} & 42.0 & \textbf{30.0} & 67.0 & \textbf{37.0} & 72.0 & 55.0 & \textbf{1.0} & 30.0 & 58.0 & 63.0 & 43.5 \\
VAoT-NoRender & 24.0 & 31.0 & \textbf{43.0} & \textbf{30.0} & \textbf{70.0} & 21.0 & 74.0 & 54.0 & 0.0 & 29.0 & 53.0 & 61.0 & 40.8 \\
VAoT & 24.0 & 31.0 & 41.0 & 19.0 & 68.0 & \textbf{37.0} & \textbf{76.0} & \textbf{59.0} & 0.0 & \textbf{32.0} & \textbf{63.0} & \textbf{75.0} & \textbf{43.8} \\
VAoT-WrongRender & 27.0 & 29.0 & 39.0 & 19.0 & 72.0 & 35.0 & 61.0 & 35.0 & 2.0 & 31.0 & 59.0 & 63.0 & 39.3 \\
\midrule
\multicolumn{14}{c}{\cellcolor{gray!12}\textit{\textbf{Gemini 3.5 Flash}}} \\
CoT & \textbf{84.0} & \textbf{74.0} & 68.0 & 15.0 & \textbf{75.0} & 55.0 & \textbf{94.0} & 61.0 & 1.0 & \textbf{42.0} & 35.0 & 80.0 & 57.0 \\
VAoT-NoRender & 77.0 & 73.0 & \textbf{70.0} & \textbf{16.0} & \textbf{75.0} & 57.0 & 92.0 & \textbf{69.0} & \textbf{3.0} & 40.0 & \textbf{39.0} & \textbf{87.0} & \textbf{58.2} \\
VAoT & 77.0 & 73.0 & 63.0 & \textbf{16.0} & \textbf{75.0} & \textbf{58.0} & 91.0 & 64.0 & 2.0 & 40.0 & 36.0 & 79.0 & 56.2 \\
VAoT-WrongRender & 76.0 & 71.0 & 63.0 & 20.0 & 73.0 & 52.0 & 75.0 & 54.0 & 4.0 & 42.0 & 33.0 & 84.0 & 53.9 \\
\midrule
\multicolumn{14}{c}{\cellcolor{gray!12}\textit{\textbf{Qwen3-VL-32B-Instruct}}} \\
CoT & 30.0 & 20.0 & \textbf{29.0} & 20.0 & \textbf{63.0} & \textbf{31.0} & 69.0 & \textbf{51.0} & 2.0 & \textbf{35.0} & \textbf{49.0} & \textbf{65.0} & \textbf{38.7} \\
VAoT-NoRender & \textbf{31.0} & 18.0 & 28.0 & \textbf{21.0} & 59.0 & 25.0 & \textbf{71.0} & 45.0 & \textbf{4.0} & \textbf{35.0} & 48.0 & 61.0 & 37.2 \\
VAoT & 30.0 & \textbf{22.0} & 26.0 & 17.0 & 58.0 & 25.0 & 65.0 & 49.0 & 2.0 & 26.0 & 46.0 & 62.0 & 35.7 \\
VAoT-WrongRender & 33.0 & 16.0 & 30.0 & 18.0 & 63.0 & 30.0 & 47.0 & 33.0 & 4.0 & 34.0 & 42.0 & 67.0 & 34.8 \\
\bottomrule
\end{tabular}
}
\end{table*}

\subsection{Experimental Setup}
\label{sec:exp_setup}

\textbf{Models and coverage.}
We evaluate GPT-5.5, GPT-o3, Gemini 3.5 Flash, and
Qwen3-VL-32B-Instruct under all four settings on the complete 1.2K-sample
benchmark. Table~\ref{tab:main_acc} and Figure~\ref{fig:task_model_patterns}
report outcome results. Table~\ref{tab:main_diagnostic_results} and
Figure~\ref{fig:process_score_patterns}(a) cover all 4,800 VAoT trajectories.
Figures~\ref{fig:process_score_patterns}(b), \ref{fig:render_transition},
and~\ref{fig:uptake_corruption} use complete sample-level alignments across
all four models. Additional statistics appear in
Appendix~\ref{app:complete_outcome_results}.

\textbf{Inference settings.}
CoT reasons textually over the original image. VAoT-NoRender generates
structured actions without executing them. VAoT executes the actions and
returns rendered states, while VAoT-WrongRender returns natural-looking,
task-relevant corrupted states as a diagnostic intervention.

\textbf{Evaluation.}
We use normalized exact matching for structured answers and a
semantic-equivalence judge for free-form predictions. For each VAoT
trajectory, GPT-5.4 selects one key visual step and scores Action Relevance,
Render Faithfulness, and Feedback Uptake on $\{0,0.5,1\}$. Model identity and
explicit correctness labels are withheld. Full prompts and validation
protocols appear in the appendix.


\subsection{Main Results}
\label{sec:main_results}

Table~\ref{tab:main_acc} reports category-level accuracy on the complete
1.2K-sample benchmark. \textbf{No single inference setting dominates}:
GPT-5.5 and Qwen3-VL-32B-Instruct perform best with CoT, GPT-o3 with VAoT,
and Gemini 3.5 Flash with VAoT-NoRender. The effect of intermediate visual
reasoning therefore depends on both the model and the task.

\textbf{Robot Manipulation is a consistent low-accuracy outlier.}
These tasks require precise target grounding and the immediate manipulation
command. Selecting a nearby object, describing the scene, or returning a
multi-step plan is counted as incorrect, leaving little room for visual actions
to compensate for grounding errors.

\begin{takeawaybox}
\textbf{Takeaway 1}\quad
Intermediate visual reasoning is a conditional capability rather than a
universal accuracy booster: the strongest setting differs across model
families.
\end{takeawaybox}


\subsection{When Does Visual Thinking Help?}
\label{sec:when_visual_thinking_helps}

\begin{figure}[t]
    \centering
    \begin{minipage}[b]{0.49\textwidth}
        \centering
        \includegraphics[width=\linewidth]{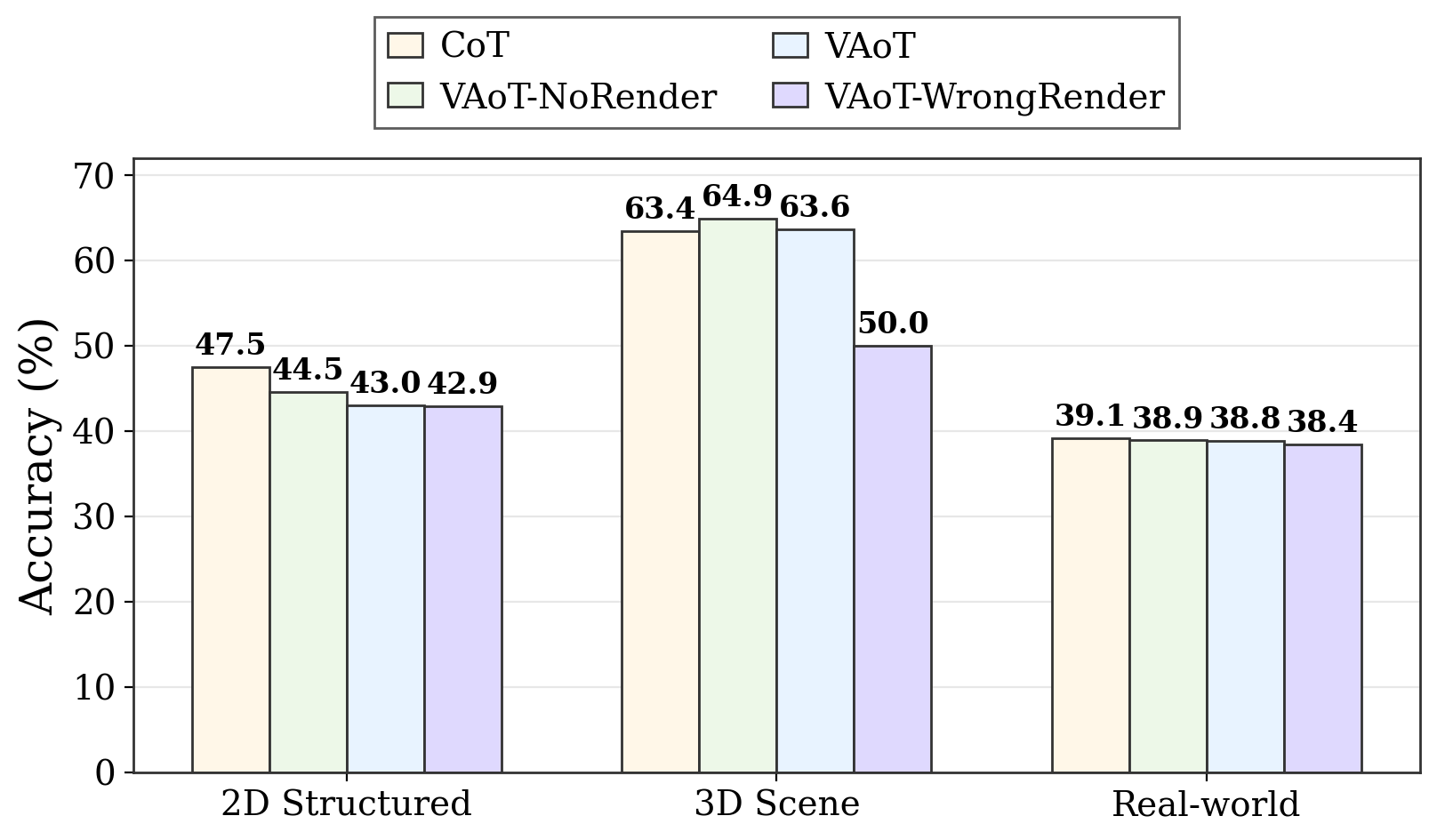}

        \vspace{2pt}
        {\small (a)}
    \end{minipage}
    \hfill
    \begin{minipage}[b]{0.49\textwidth}
        \centering
        \includegraphics[width=\linewidth]{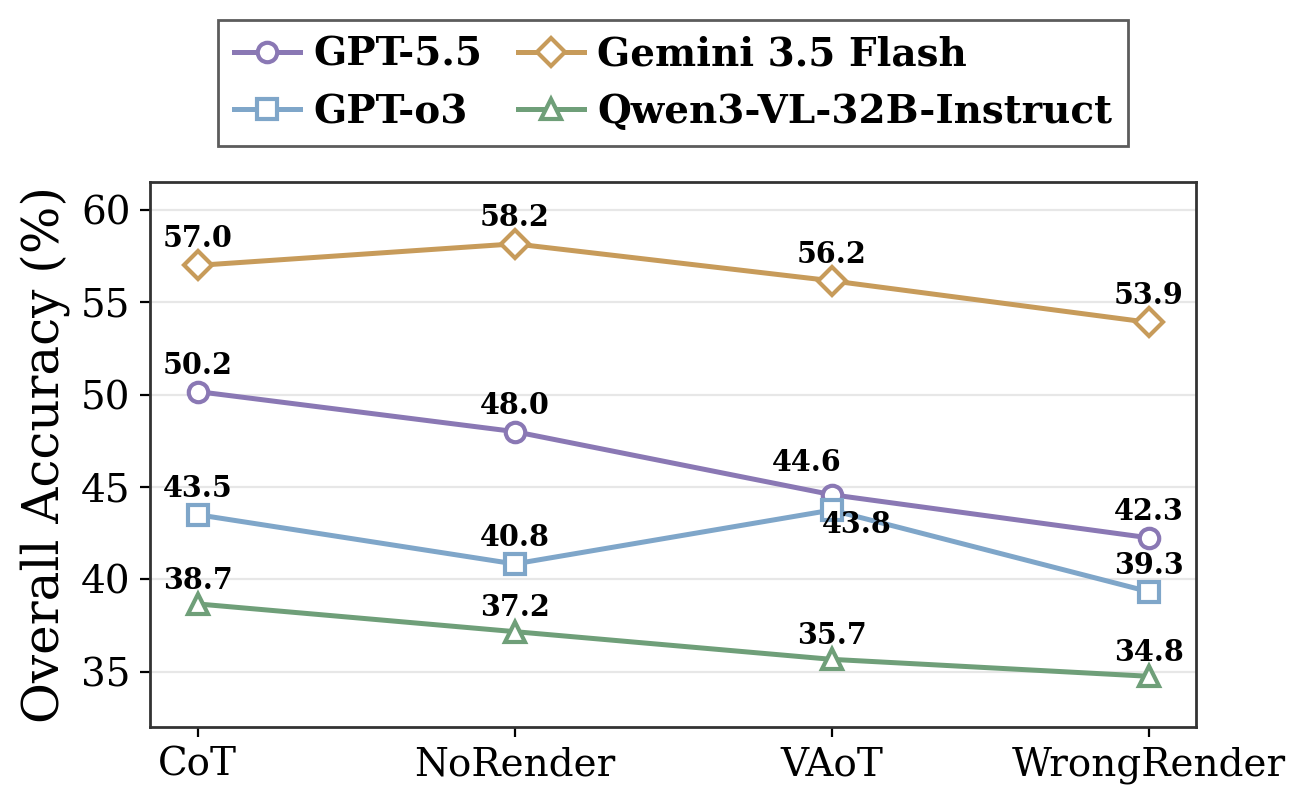}

        \vspace{2pt}
        {\small (b)}
    \end{minipage}
    \caption{\textbf{Visual-action reasoning across task groups and models.}
(a) Accuracy under the four inference settings, averaged across models within
each task group. (b) Overall accuracy for each model and setting. All results
use the complete 1.2K-sample benchmark.}
    \label{fig:task_model_patterns}
\end{figure}

Figure~\ref{fig:task_model_patterns} summarizes the results by broad task group
and by model.

\textbf{Task regimes.}
Structured 2D problems favor direct reasoning over the original image,
especially for geometry and visual puzzles. In 3D scenes, VAoT-NoRender is
strongest on average, indicating that explicitly selecting objects and
relations can organize relational reasoning even before an action is rendered.
In real-world scenes, CoT, VAoT-NoRender, and VAoT are nearly tied on average,
showing no clear aggregate advantage for a single reasoning strategy.

\textbf{Model profiles.}
GPT-5.5 and Qwen3-VL-32B-Instruct exhibit \textit{text-first} profiles,
Gemini 3.5 Flash a \textit{planner-first} profile, and GPT-o3 a slight
\textit{closed-loop} preference. Thus, models differ not only in final accuracy
but also in the stage of visual reasoning from which they benefit.

\begin{takeawaybox}
\textbf{Takeaway 2}\quad
Visual thinking changes regime across environments: direct perception leads
on explicit 2D diagrams, action planning slightly leads in 3D scenes, and no
single strategy has a clear aggregate advantage in real-world environments.
\end{takeawaybox}

VAoT-WrongRender generally lowers accuracy, including cases in which
VAoT does not outperform VAoT-NoRender. A model can therefore depend on
returned visual feedback without obtaining a net gain from correct rendering.

\begin{takeawaybox}
\textbf{Takeaway 3}\quad
Visual-state utility and visual-state dependence are not equivalent:
correct rendering may yield little net gain, while corrupting the returned
state can still cause measurable performance degradation.
\end{takeawaybox}


\subsection{Explaining Outcome Patterns through Process-level Diagnosis}
\label{sec:process_diagnosis}

To localize the source of the outcome differences, we decompose VAoT
into three stages: \textit{planning}, \textit{rendering}, and
\textit{feedback use}.
Action Relevance measures whether the requested operation targets relevant
visual evidence; Render Faithfulness measures whether the operation is
correctly realized; Feedback Uptake measures whether subsequent reasoning uses
information from the returned visual state.

\begin{table*}[t]
\centering
\small
\caption{\textbf{Process-level diagnosis on the complete VAoT trajectories.}
Average Action Relevance, Render Faithfulness, and Feedback Uptake across the
three visual-reasoning groups and overall. Each dimension is scored on
$\{0,0.5,1\}$ for one key visual step per trajectory. Each model contributes
1,200 trajectories.}
\label{tab:main_diagnostic_results}
\begin{tabular*}{\textwidth}{@{\extracolsep{\fill}}llcccc@{}}
\toprule
\textbf{Model}
& \textbf{Metric}
& \textbf{2D Structured}
& \textbf{3D Scene}
& \textbf{Real-world}
& \textbf{Overall} \\
\midrule
\multirow{3}{*}{GPT-5.5}
& Action Relevance    & 0.980 & 0.995 & 0.988 & 0.985 \\
& Render Faithfulness & 0.595 & 0.550 & 0.681 & 0.616 \\
& Feedback Uptake     & 0.723 & 0.785 & 0.837 & 0.772 \\
\midrule
\multirow{3}{*}{GPT-o3}
& Action Relevance    & 0.994 & 0.992 & 0.948 & 0.978 \\
& Render Faithfulness & 0.588 & 0.593 & 0.664 & 0.614 \\
& Feedback Uptake     & 0.812 & 0.843 & 0.852 & 0.830 \\
\midrule
\multirow{3}{*}{Gemini 3.5 Flash}
& Action Relevance    & 0.974 & 0.988 & 0.972 & 0.976 \\
& Render Faithfulness & 0.599 & 0.705 & 0.639 & 0.630 \\
& Feedback Uptake     & 0.656 & 0.812 & 0.691 & 0.694 \\
\midrule
\multirow{3}{*}{Qwen3-VL-32B-Instruct}
& Action Relevance    & 0.958 & 0.978 & 0.949 & 0.958 \\
& Render Faithfulness & 0.563 & 0.593 & 0.641 & 0.594 \\
& Feedback Uptake     & 0.832 & 0.913 & 0.909 & 0.871 \\
\bottomrule
\end{tabular*}
\end{table*}

Action Relevance is consistently near saturation, whereas Render Faithfulness
is substantially lower across all models and task groups. \textbf{The largest
observed degradation therefore occurs after action selection}: a plausible visual plan
must still be rendered accurately and incorporated into later reasoning.
Qwen3-VL-32B-Instruct has the highest overall Feedback Uptake (0.871)
despite lower final-answer accuracy, further illustrating that using returned
visual information is distinct from using it correctly. The human audit in
Section~\ref{sec:human_alignment} evaluates the reliability of these automatic
judgments.

\begin{figure*}[t]
\centering
\begin{minipage}[b]{0.52\textwidth}
    \centering
    \makebox[\linewidth][c]{%
        \includegraphics[height=4.35cm,keepaspectratio]{%
            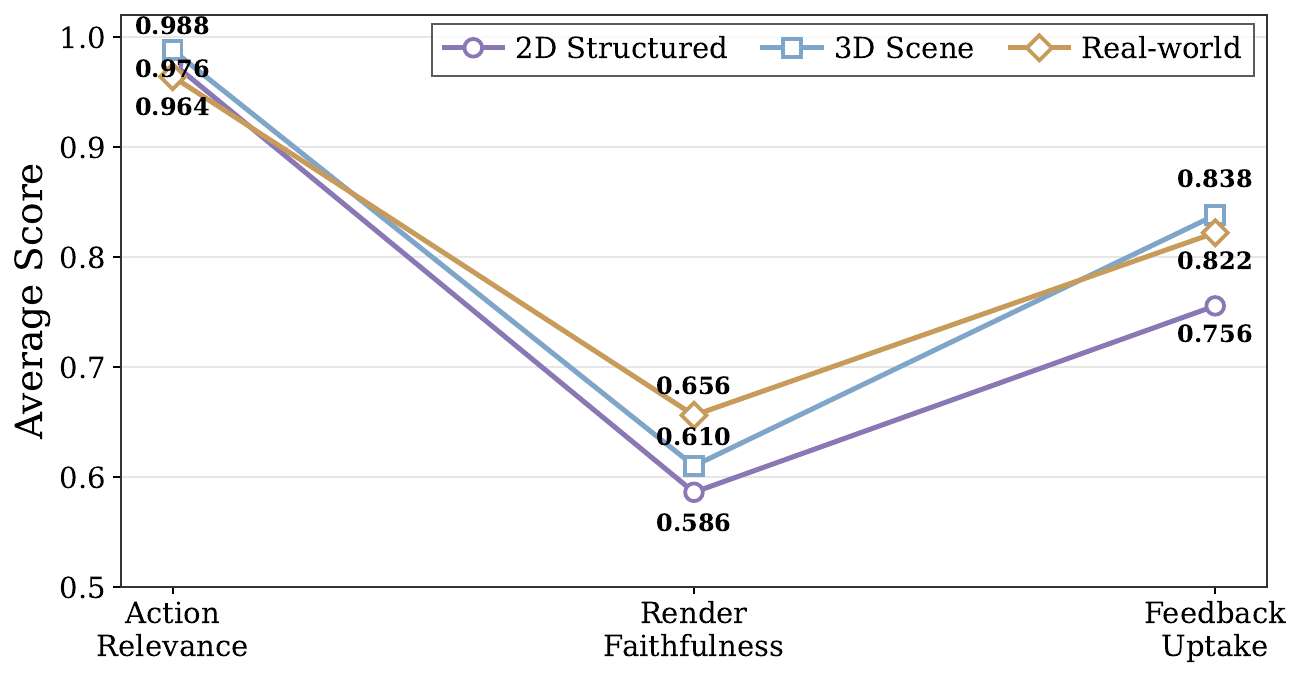}}

    \vspace{2pt}
    {\small (a)}
\end{minipage}
\hfill
\begin{minipage}[b]{0.42\textwidth}
    \centering
    \makebox[\linewidth][c]{%
        \includegraphics[height=4.35cm,keepaspectratio]{%
            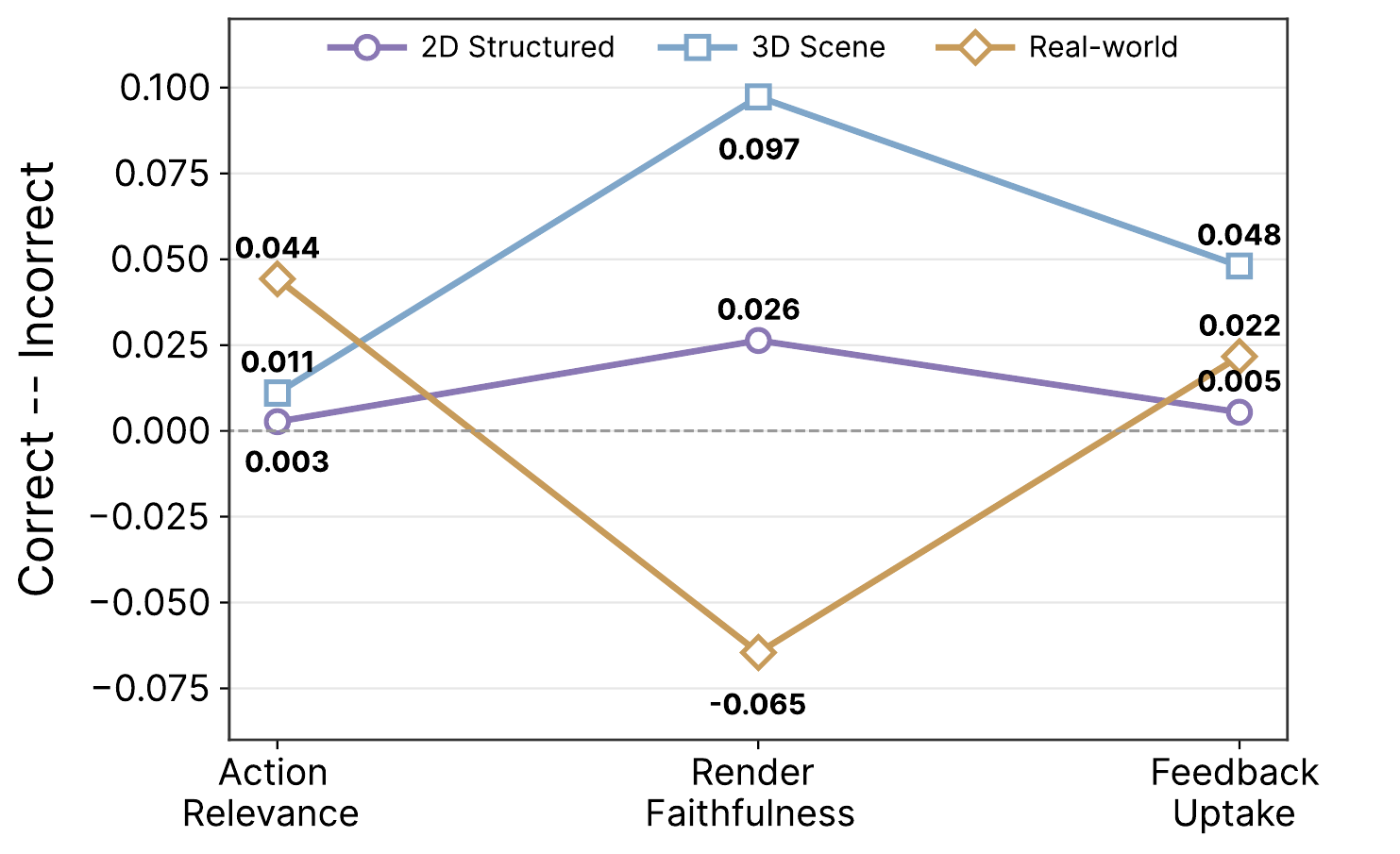}}

    \vspace{2pt}
    {\small (b)}
\end{minipage}

\caption{\textbf{Process-level diagnosis across visual environments.}
(a) Average Action Relevance, Render Faithfulness, and Feedback Uptake across
the three task groups, averaged over all four models.
(b) Outcome-stratified correct-minus-incorrect differences in the three
process scores, averaged over the four models.}
\label{fig:process_score_patterns}
\end{figure*}

\textbf{Different visual environments fail at different stages.}
Figure~\ref{fig:process_score_patterns}(b) provides an outcome-stratified
process analysis averaged over all four models. 3D scene reasoning has a
pronounced correct--incorrect gap in Render Faithfulness (0.097),
substantially larger than its Action Relevance (0.011) and Feedback Uptake
(0.048) gaps. Real-world reasoning instead separates successful trajectories
most clearly through Action Relevance (0.044), while its Render Faithfulness
gap is negative ($-0.065$) and its Feedback Uptake gap is smaller (0.022). For
2D tasks, all three gaps remain small (at most 0.026), consistent with the
original diagrams already exposing much of the required structure.

\begin{takeawaybox}
\textbf{Takeaway 4}\quad
The main bottleneck is not selecting a task-relevant visual action.
It lies in faithfully executing that action and using the resulting evidence,
with the dominant failure stage varying across visual environments.
\end{takeawaybox}

Complete correctness-split statistics are reported in
Appendix~\ref{app:complete_process_statistics}.

\textbf{Rendering benefit and rendering harm.}
To separate opposing effects of rendered feedback, we compare aligned
VAoT-NoRender and VAoT trajectories within each Render Faithfulness group.
\textit{Render benefit} is an incorrect-to-correct transition, and
\textit{render harm} is the reverse transition.

\begin{figure*}[t]
\centering
\includegraphics[width=0.8\textwidth]{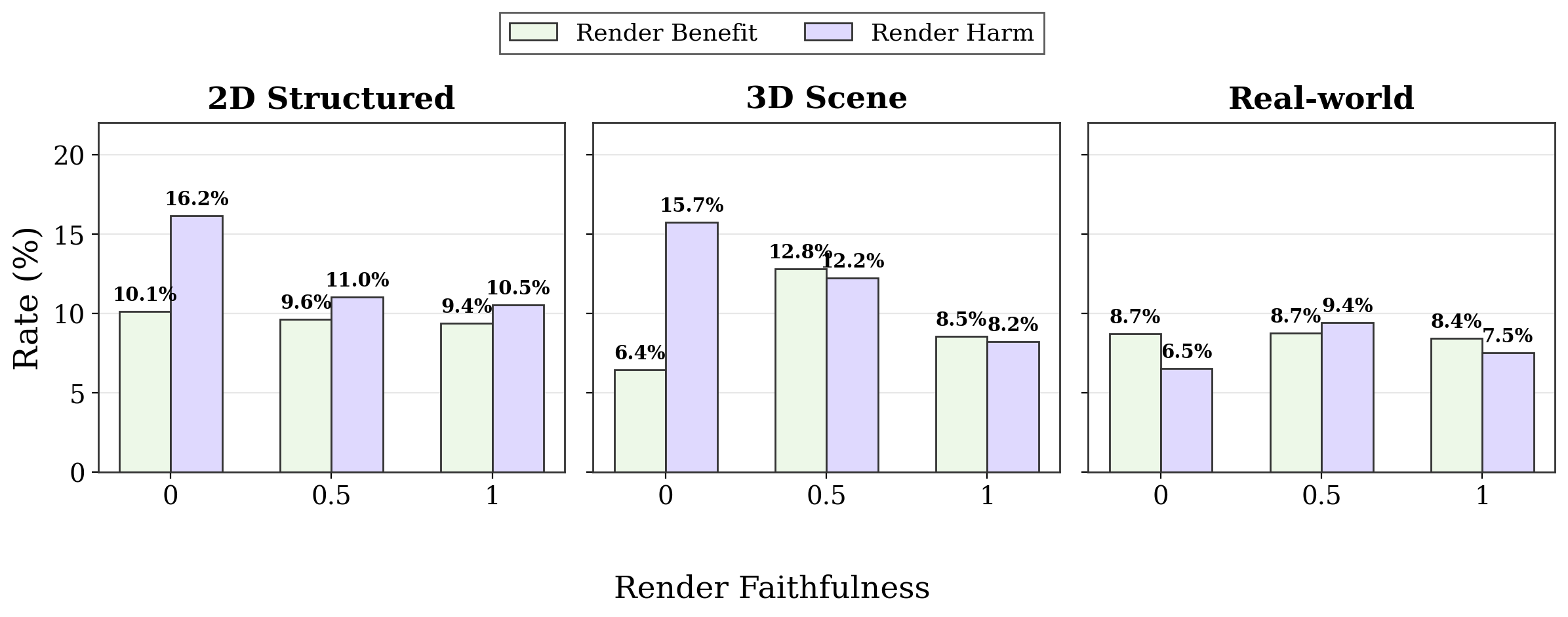}
\caption{\textbf{Rendering benefit and harm by Render Faithfulness.}
Benefit denotes a VAoT-NoRender incorrect-to-VAoT correct transition,
and harm denotes the reverse. Rates are normalized within each task group
and Render Faithfulness bin for the four-model aligned subset.}
\label{fig:render_transition}
\vspace{-20pt}
\end{figure*}

On the four-model aligned subset,
Figure~\ref{fig:render_transition} shows that completely unfaithful
rendering produces a negative transition balance in 2D and 3D reasoning
($-6.1$ and $-9.3$ percentage points, respectively), while the real-world
group is slightly positive ($+2.2$). Higher Render Faithfulness primarily
reduces render harm. At score $1$, the transition balance remains slightly
negative in 2D ($-1.2$ points), but becomes mildly positive in 3D ($+0.3$)
and real-world reasoning ($+0.9$).

\begin{takeawaybox}
\textbf{Takeaway 5}\quad
Render Faithfulness mainly determines whether execution preserves or damages
the planning gain: unfaithful renders are especially harmful in 2D and 3D
reasoning, while more faithful renders mostly reduce disruption and shift the
net balance toward neutral or mildly positive outcomes.
\end{takeawaybox}

\noindent
\begin{minipage}[t]{0.48\textwidth}
    \vspace{0pt}
    \small
    \raggedright
    \textbf{Feedback uptake and behavioral dependence.}
    To test whether Feedback Uptake corresponds to observable dependence on
    the returned visual state, we align VAoT and VAoT-WrongRender
    trajectories and group them by the Feedback Uptake score assigned to the
    VAoT trajectory. We report
    $\mathrm{Acc}_{\mathrm{VAoT}}-\mathrm{Acc}_{\mathrm{WrongRender}}$,
    where larger positive values indicate greater sensitivity to corrupted
    feedback.

    \vspace{5pt}
    Overall accuracy degradation changes from $-2.3$ to $3.4$ and $3.1$
    percentage points as Feedback Uptake rises from $0$ to $0.5$ and $1$.
    \textbf{The relationship is strongest in 3D scene reasoning}, where the drop
    grows from $3.5$ to $10.3$ and $15.5$ points.
\end{minipage}
\hfill
\begin{minipage}[t]{0.4\textwidth}
    \vspace{0pt}
    \captionsetup{type=figure}
    \captionof{figure}{\textbf{Accuracy degradation under corrupted feedback.}
    Results use complete sample-level alignments across all 1.2K samples for
    all four models; larger values indicate greater sensitivity.}
    \label{fig:uptake_corruption}

    \vspace{2pt}
    \centering
    \includegraphics[width=1\linewidth]{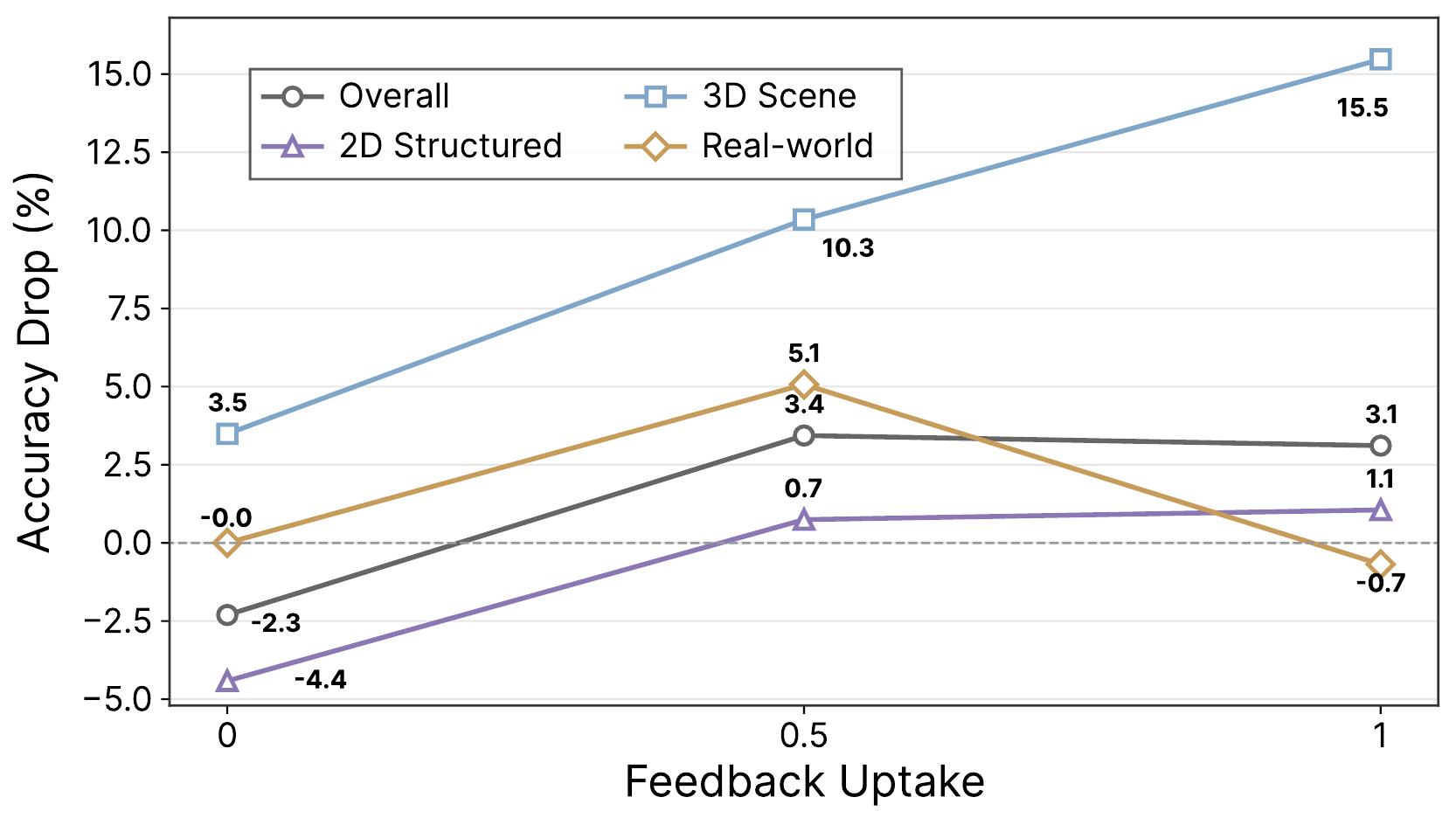}
\end{minipage}
\par\smallskip

The association is weaker or non-monotonic in 2D and real-world tasks,
indicating that the behavioral validity of Feedback Uptake remains
environment-dependent. Nevertheless, corrupted feedback changes the semantic
answer in 32.7--54.2\% of samples across the four models, even when its net
accuracy effect is smaller. Complete model-level counts are reported in
Appendix~\ref{tab:semantic_answer_change}.

\medskip

\begin{takeawaybox}
\textbf{Takeaway 6}\quad
Feedback Uptake tracks corrupted-feedback sensitivity most clearly in 3D
reasoning, but the association remains environment-dependent.
\end{takeawaybox}


\subsection{Human Validation of Process-level Evaluation}
\label{sec:human_alignment}

\noindent
\begin{minipage}[t]{0.46\textwidth}
    \vspace{0pt}
    Because the process analysis relies on an external multimodal judge, we
    conduct a diagnostically stratified human audit. Two human annotators
    review 240 VAoT trajectories drawn from a
    $4\times3\times4\times5$ design over GPT-5.5, GPT-o3,
    Gemini 3.5 Flash, and Qwen3-VL-32B-Instruct, three task groups, four
    diagnostic strata, and five samples per stratum.

    \vspace{5pt}
    For each trajectory, the annotators independently assess the selected key
    step and the three process scores as \textit{reasonable},
    \textit{partially reasonable}, or \textit{unreasonable}. Pooling their 480
    ratings per target, 92.9--96.9\% of judgments are at least partially
    reasonable across all four targets.
\end{minipage}
\hfill
\begin{minipage}[t]{0.50\textwidth}
    \vspace{0pt}
    \captionsetup{type=table}
    \captionof{table}{\textbf{Human alignment of process-level evaluation.}
    Rates are pooled across two independent annotators (480 ratings per
    target).}
    \label{tab:human_alignment}

    \vspace{3pt}
    \centering
    \small
    \setlength{\tabcolsep}{5pt}
    \renewcommand{\arraystretch}{1.08}
    \begin{tabular}{@{}lcc@{}}
    \toprule
    \textbf{Annotation target}
    & \textbf{Reasonable}
    & \shortstack{\textbf{Reasonable or}\\\textbf{partially reasonable}} \\
    \midrule
    Key-step selection   & 88.3\% & 94.4\% \\
    Action relevance     & 74.8\% & 96.9\% \\
    Render faithfulness  & 70.2\% & 92.9\% \\
    Feedback uptake      & 79.6\% & 95.0\% \\
    \bottomrule
    \end{tabular}
\end{minipage}
\par\medskip

The larger strict-to-partial gaps for Action Relevance and Render Faithfulness
indicate disagreement mainly at the boundary between fully and partially
supported judgments; pooled reasonable-or-partial rates exceed 92.9\% for every
target. Sampling, annotation details, and per-annotator human-validation
results are provided in Appendix~\ref{app:human_process_validation}.
WrongRender quality-audit results and paired accuracy analyses on
quality-filtered subsets are provided in Appendix~\ref{app:wrongrender_audit}. The VAoT--WrongRender accuracy gap remains positive under both the strict and relaxed quality criteria, although its magnitude varies with intervention quality. Representative successful, diagnostic, and corrupted-feedback trajectories are shown in Appendix~\ref{app:qualitative_cases}.

\section{Related Work}
\label{sec:related_work}

\textbf{Multimodal chain-of-thought and active visual reasoning.}
Textual chain-of-thought has been extended from language reasoning to
multimodal problem solving, with evaluations covering rationale quality,
robustness, efficiency, and error localization
\citep{zhang2023multimodalcot,jiang2025mmecot}.
A rapidly growing line of work instead treats vision as an active reasoning
workspace. Tool-based methods crop, zoom, annotate, sketch, or otherwise
manipulate task-relevant image regions
\citep{hu2024visualsketchpad,zheng2025deepeyes,
su2025openthinkimg}.
Interactive and generative approaches construct intermediate visual states
during reasoning through image editing, visual generation, or interleaved
text--visual inference
\citep{shi2025mathcanvas,qiao2025vthinker,
liu2026dapicot,li2026s1vl}.
Recent work further explores continuous visual actions and repeated access to
visual information during multimodal reasoning
\citep{zhao2026nvcot,hu2026tvicot}.
Reliability-oriented approaches address noisy intermediate images or
discrepancies between textual plans and executed visual actions
\citep{li2026reliablethinking,yang2026walkthetalk}.
These methods demonstrate the potential of active visual reasoning, but
performance improvements alone do not establish that a returned visual state
faithfully realizes the intended action or is subsequently used by the model.

\textbf{Benchmarks for visual intermediate states.}
MIRA studies tasks in which intermediate visualizations can facilitate
reasoning, while ViC-Bench evaluates free-form visual-interleaved reasoning
trajectories
\citep{zhou2025mira,wu2025vicbench}.
TWI-PRMBench focuses on process reward modeling for thinking-with-images
trajectories, and TwiFF-Bench evaluates dynamically generated future-frame
reasoning
\citep{zhou2026thinkwithimages,liu2026twiff}.
These benchmarks substantially broaden the evaluation of intermediate visual
reasoning. However, they primarily measure the usefulness, plausibility, or
overall quality of visual traces. They do not jointly diagnose whether a
self-generated visual action is task-relevant, whether the returned state
faithfully realizes that action, and whether later reasoning behaviorally
depends on the returned visual evidence.

\textbf{Diagnosing visual-state use.}
Several recent studies move closer to evaluating whether intermediate visual
information is faithful or answer-critical. Faithfulness-oriented evaluation
examines whether generated visual thoughts are consistent with the intended
reasoning process
\citep{liu2025faithfulvisualthinking}.
View Dropout encourages unified multimodal models to rely on generated
thinking images by restricting direct information flow from the original
views to the final answer
\citep{yang2026howwhatimagine}.
Systematic comparisons of tool-based and tool-free agents further show that
producing tool-call traces does not necessarily imply that the returned tool
outputs expand the set of problems an agent can solve
\citep{guo2026toolbenefit}.
See2Think is complementary to these studies. Rather than introducing a
training objective or comparing separately trained tool and non-tool agents,
it provides a model-agnostic inference-time framework that records
self-generated visual actions, externally rendered states, and subsequent
reasoning. Matched standard and corrupted-render conditions then distinguish
action relevance, render faithfulness, feedback uptake, visual-state utility,
and behavioral dependence.

\section{Conclusion and Limitations}
\label{sec:conclusion}

We introduced \textbf{See2Think}, a unified framework for studying whether
multimodal models genuinely use intermediate visual states during reasoning.
See2Think combines \textbf{See2ThinkBench}, a benchmark of visually
dependent reasoning problems evaluated through a unified answer-generation
interface, and \textbf{VAoT}, an observable and intervenable protocol that records
visual actions, rendered states, and subsequent reasoning. Experiments across GPT-5.5, GPT-o3, Gemini 3.5 Flash, and Qwen3-VL-32B-Instruct reveal that
visual reasoning is model- and environment-dependent rather than governed by a
single universally effective strategy. While models generally identify
relevant visual operations, faithfully realizing these operations and using
the returned visual evidence remain important bottlenecks. Controlled
interventions further show that visual-state utility and behavioral dependence
are distinct: a visual state may provide limited direct benefit while still
strongly affecting model decisions when corrupted. By moving beyond the mere
existence of visual traces and final-answer accuracy, See2Think provides a
controlled framework for evaluating when ``thinking with images'' is genuine,
useful, or misleading.

\textbf{Limitations.}
First, our evaluation focuses on four representative multimodal models and may not capture the full diversity of current and future vision-language systems. Second, VAoT relies on an external rendering component and automatic process evaluation, which may introduce
errors despite our human validation. Third, the measured effect of VAoT-WrongRender depends on intervention quality. Quality-filtered analyses on the human-audited subset preserve the
direction of the accuracy drop, but its magnitude varies across filtering criteria.
Future work may explore broader model coverage, learned rendering mechanisms,
and more fine-grained process supervision for visual reasoning.

\bibliographystyle{plainnat}
\bibliography{see2think_related_refs}

@inproceedings{wei2022chain,
  title={Chain-of-Thought Prompting Elicits Reasoning in Large Language Models},
  author={Wei, Jason and Wang, Xuezhi and Schuurmans, Dale and Bosma, Maarten and Ichter, Brian and Xia, Fei and Chi, Ed and Le, Quoc V. and Zhou, Denny},
  booktitle={Advances in Neural Information Processing Systems},
  year={2022}
}

@article{zhang2023multimodalcot,
  title={Multimodal Chain-of-Thought Reasoning in Language Models},
  author={Zhang, Zhuosheng and Zhang, Aston and Li, Mu and Zhao, Hai and Karypis, George and Smola, Alexander J.},
  journal={arXiv preprint arXiv:2302.00923},
  year={2023}
}

@inproceedings{jiang2025mmecot,
  title={{MME}-{C}o{T}: Benchmarking Chain-of-Thought in Large Multimodal Models for Reasoning Quality, Robustness, and Efficiency},
  author={Jiang, Dongzhi and Zhang, Renrui and Guo, Ziyu and Li, Yanwei and Qi, Yu and Chen, Xinyan and Wang, Liuhui and Jin, Jianhan and Guo, Claire and Yan, Shen and Zhang, Bo and Fu, Chaoyou and Gao, Peng and Li, Hongsheng},
  booktitle={Proceedings of the 42nd International Conference on Machine Learning},
  pages={27793--27830},
  year={2025},
  volume={267},
  series={Proceedings of Machine Learning Research},
  publisher={PMLR}
}

@article{hu2024visualsketchpad,
  title={Visual Sketchpad: Sketching as a Visual Chain of Thought for Multimodal Language Models},
  author={Hu, Yushi and Shi, Weijia and Fu, Xingyu and Roth, Dan and Ostendorf, Mari and Zettlemoyer, Luke and Smith, Noah A. and Krishna, Ranjay},
  journal={arXiv preprint arXiv:2406.09403},
  year={2024}
}

@article{zheng2025deepeyes,
  title={DeepEyes: Incentivizing ``Thinking with Images'' via Reinforcement Learning},
  author={Zheng, Ziwei and Yang, Michael and Hong, Jack and Zhao, Chenxiao and Xu, Guohai and Yang, Le and Shen, Chao and Yu, Xing},
  journal={arXiv preprint arXiv:2505.14362},
  year={2025}
}

@article{su2025openthinkimg,
  title={OpenThinkIMG: Learning to Think with Images via Visual Tool Reinforcement Learning},
  author={Su, Zhaochen and Li, Linjie and Song, Mingyang and Hao, Yunzhuo and Yang, Zhengyuan and Zhang, Jun and Chen, Guanjie and Gu, Jiawei and Li, Juntao and Qu, Xiaoye and Cheng, Yu},
  journal={arXiv preprint arXiv:2505.08617},
  year={2025}
}

@article{shi2025mathcanvas,
  title={MathCanvas: Intrinsic Visual Chain-of-Thought for Multimodal Mathematical Reasoning},
  author={Shi, Weikang and Yu, Aldrich and Fang, Rongyao and Ren, Houxing and Wang, Ke and Zhou, Aojun and Tian, Changyao and Fu, Xinyu and Hu, Yuxuan and Lu, Zimu and Huang, Linjiang and Liu, Si and Liu, Rui and Li, Hongsheng},
  journal={arXiv preprint arXiv:2510.14958},
  year={2025}
}

@article{qiao2025vthinker,
  title={V-Thinker: Interactive Thinking with Images},
  author={Qiao, Runqi and Tan, Qiuna and Yang, Minghan and Dong, Guanting and Yang, Peiqing and Lang, Shiqiang and Wan, Enhui and Wang, Xiaowan and Xu, Yida and Yang, Lan and Sun, Chong and Li, Chen and Zhang, Honggang},
  journal={arXiv preprint arXiv:2511.04460},
  year={2025}
}

@article{zhou2025mira,
  title={When Visualizing is the First Step to Reasoning: MIRA, a Benchmark for Visual Chain-of-Thought},
  author={Zhou, Yiyang and Tu, Haoqin and Wang, Zijun and Wang, Zeyu and Muennighoff, Niklas and Nie, Fan and Choi, Yejin and Zou, James and Deng, Chaorui and Yan, Shen and Fan, Haoqi and Xie, Cihang and Yao, Huaxiu and Ye, Qinghao},
  journal={arXiv preprint arXiv:2511.02779},
  year={2025}
}

@misc{rbenchv,
  title         = {RBench-V: A Primary Assessment for Visual Reasoning Models with Multi-modal Outputs},
  author        = {Guo, Meng-Hao and Chu, Xuanyu and Yang, Qianrui and Mo, Zhe-Han and Shen, Yiqing and Li, Pei-lin and Lin, Xinjie and Zhang, Jinnian and Chen, Xin-Sheng and Zhang, Yi and Nakayama, Kiyohiro and Geng, Zhengyang and Peng, Houwen and Hu, Han and Hu, Shi-Min},
  year          = {2025},
  eprint        = {2505.16770},
  archivePrefix = {arXiv},
  primaryClass  = {cs.CV}
}

@misc{emma,
  title         = {Can MLLMs Reason in Multimodality? EMMA: An Enhanced MultiModal ReAsoning Benchmark},
  author        = {Hao, Yunzhuo and Gu, Jiawei and Wang, Huichen Will and Li, Linjie and Yang, Zhengyuan and Wang, Lijuan and Cheng, Yu},
  year          = {2025},
  eprint        = {2501.05444},
  archivePrefix = {arXiv},
  primaryClass  = {cs.CV}
}

@inproceedings{m3cot,
  title     = {{M}$^3${C}o{T}: A Novel Benchmark for Multi-Domain Multi-step Multi-modal Chain-of-Thought},
  author    = {Chen, Qiguang and Qin, Libo and Zhang, Jin and Chen, Zhi and Xu, Xiao and Che, Wanxiang},
  booktitle = {Proceedings of the 62nd Annual Meeting of the Association for Computational Linguistics (Volume 1: Long Papers)},
  year      = {2024},
  pages     = {8199--8221},
  address   = {Bangkok, Thailand},
  publisher = {Association for Computational Linguistics},
  doi       = {10.18653/v1/2024.acl-long.446},
  url       = {https://aclanthology.org/2024.acl-long.446/}
}

@misc{prismbench,
  title         = {PRISM-Bench: A Benchmark of Puzzle-Based Visual Tasks with CoT Error Detection},
  author        = {Qian, Yusu and Wan, Cheng and Jia, Chao and Yang, Yinfei and Zhao, Qingyu and Gan, Zhe},
  year          = {2025},
  eprint        = {2510.23594},
  archivePrefix = {arXiv},
  primaryClass  = {cs.CV},
  note          = {Withdrawn}
}

@inproceedings{johnson2017clevr,
  title     = {{CLEVR}: A Diagnostic Dataset for Compositional Language and Elementary Visual Reasoning},
  author    = {Johnson, Justin and Hariharan, Bharath and van der Maaten, Laurens and Fei-Fei, Li and Zitnick, C. Lawrence and Girshick, Ross},
  booktitle = {Proceedings of the IEEE Conference on Computer Vision and Pattern Recognition},
  year      = {2017},
  pages     = {1988--1997}
}

@inproceedings{superclevr,
  title     = {Super-{CLEVR}: A Virtual Benchmark to Diagnose Domain Robustness in Visual Reasoning},
  author    = {Li, Zhuowan and Wang, Xingrui and Stengel-Eskin, Elias and Kortylewski, Adam and Ma, Wufei and Van Durme, Benjamin and Yuille, Alan},
  booktitle = {Proceedings of the IEEE/CVF Conference on Computer Vision and Pattern Recognition},
  year      = {2023}
}

@misc{vlabench,
  title         = {{VLABench}: A Large-Scale Benchmark for Language-Conditioned Robotics Manipulation with Long-Horizon Reasoning Tasks},
  author        = {Zhang, Shiduo and Xu, Zhe and Liu, Peiju and Yu, Xiaopeng and Li, Yuan and Gao, Qinghui and Fei, Zhaoye and Yin, Zhangyue and Wu, Zuxuan and Jiang, Yu-Gang and Qiu, Xipeng},
  year          = {2024},
  eprint        = {2412.18194},
  archivePrefix = {arXiv},
  primaryClass  = {cs.RO}
}

@misc{droid,
  title         = {{DROID}: A Large-Scale In-The-Wild Robot Manipulation Dataset},
  author        = {Khazatsky, Alexander and Pertsch, Karl and Nair, Suraj and Balakrishna, Ashwin and Dasari, Sudeep and Karamcheti, Siddharth and Nasiriany, Soroush and Srirama, Mohan Kumar and Chen, Lawrence Yunliang and Ellis, Kirsty and others},
  year          = {2024},
  eprint        = {2403.12945},
  archivePrefix = {arXiv},
  primaryClass  = {cs.RO}
}

@misc{intphys2,
  title         = {IntPhys 2: Benchmarking Intuitive Physics Understanding In Complex Synthetic Environments},
  author        = {Bordes, Florian and Garrido, Quentin and Kao, Justine T. and Williams, Adina and Rabbat, Michael and Dupoux, Emmanuel},
  year          = {2025},
  eprint        = {2506.09849},
  archivePrefix = {arXiv},
  primaryClass  = {cs.CV}
}

@inproceedings{lu2023mathvista,
  title     = {{MathVista}: Evaluating Mathematical Reasoning of Foundation Models in Visual Contexts},
  author    = {Lu, Pan and Bansal, Hritik and Xia, Tony and Liu, Jiacheng and Li, Chunyuan and Hajishirzi, Hannaneh and Cheng, Hao and Chang, Kai-Wei and Galley, Michel and Gao, Jianfeng},
  booktitle = {International Conference on Learning Representations},
  year      = {2024}
}

@inproceedings{yue2024mmmu,
  title     = {{MMMU}: A Massive Multi-discipline Multimodal Understanding and Reasoning Benchmark for Expert {AGI}},
  author    = {Yue, Xiang and Ni, Yuansheng and Zhang, Kai and Zheng, Tianyu and Liu, Ruoqi and Zhang, Ge and Stevens, Samuel and Jiang, Dongfu and Ren, Weiming and Sun, Yuxuan and Wei, Cong and Yu, Botao and Yuan, Ruibin and Sun, Renliang and Yin, Ming and Zheng, Boyuan and Yang, Zhenzhu and Liu, Yibo and Huang, Wenhao and Sun, Huan and Su, Yu and Chen, Wenhu},
  booktitle = {Proceedings of the IEEE/CVF Conference on Computer Vision and Pattern Recognition},
  year      = {2024}
}

@article{wu2025vicbench,
title={ViC-Bench: Benchmarking Visual-Interleaved Chain-of-Thought Capability in MLLMs with Free-Style Intermediate State Representations},
author={Wu, Xuecheng and Liu, Jiaxing and Huang, Danlei and Li, Xiaoyu and Wang, Yifan and Chen, Chen and Ma, Liya and Cao, Xuezhi and Xue, Junxiao},
journal={arXiv preprint arXiv:2505.14404},
year={2025}
}

@article{zhou2026thinkwithimages,
title={What, Whether and How? Unveiling Process Reward Models for Thinking with Images Reasoning},
author={Zhou, Yujin and Wen, Pengcheng and Chen, Jiale and Yin, Boqin and Zhu, Han and Ji, Jiaming and Dai, Juntao and Chan, Chi-Min and Han, Sirui},
journal={arXiv preprint arXiv:2602.08346},
year={2026}
}

@article{liu2025faithfulvisualthinking,
  title={On the Faithfulness of Visual Thinking: Measurement and Enhancement},
  author={Liu, Zujing and Pan, Junwen and She, Qi and Gao, Yuan and Xia, Guisong},
  journal={arXiv preprint arXiv:2510.23482},
  year={2025}
}

@article{liu2026twiff,
  title={{TwiFF} (Think With Future Frames): A Large-Scale Dataset for Dynamic Visual Reasoning},
  author={Liu, Junhua and Wang, Zhangcheng and Han, Zhike and Wang, Ningli and Liang, Guotao and Kuang, Kun},
  journal={arXiv preprint arXiv:2602.10675},
  year={2026}
}

@article{li2026reliablethinking,
  title={Reliable Thinking with Images},
  author={Li, Haobin and Yang, Yutong and Lin, Yijie and Dai, Xiang and Yang, Mouxing and Peng, Xi},
  journal={arXiv preprint arXiv:2602.12916},
  year={2026}
}

@article{zhao2026nvcot,
  title={Thinking with Images as Continuous Actions: Numerical Visual Chain-of-Thought},
  author={Zhao, Kesen and Zhu, Beier and Zhou, Junbao and Zhu, Xingyu and Yue, Zhongqi and Zhang, Hanwang},
  journal={arXiv preprint arXiv:2602.23959},
  year={2026}
}

@article{liu2026dapicot,
  title={Let's Think with Images Efficiently! An Interleaved-Modal Chain-of-Thought Reasoning Framework with Dynamic and Precise Visual Thoughts},
  author={Liu, Xu and Zhang, Yongheng and Chen, Qiguang and Li, Yao and Wang, Sheng and Qin, Libo},
  journal={arXiv preprint arXiv:2603.21754},
  year={2026}
}

@article{yang2026walkthetalk,
  title={Walk the Talk: Bridging the Reasoning-Action Gap for Thinking with Images via Multimodal Agentic Policy Optimization},
  author={Yang, Wenhao and Xia, Yu and Huang, Jinlong and Lu, Shiyin and Chen, Qing-Guo and Xu, Zhao and Luo, Weihua and Zhang, Kaifu and Zhou, Yuchen and Xia, Xiaobo and Wan, Yuanyu and Zhang, Lijun and Chua, Tat-Seng},
  journal={arXiv preprint arXiv:2604.06777},
  year={2026}
}

@article{li2026s1vl,
  title={{S1-VL}: Scientific Multimodal Reasoning Model with Thinking-with-Images},
  author={Li, Qingxiao and Xu, Lifeng and Wang, QingLi and Bai, Yudong and Ou, Mingwei and Hu, Shu and Xu, Nan},
  journal={arXiv preprint arXiv:2604.21409},
  year={2026}
}

@article{yang2026howwhatimagine,
  title={How and What to Imagine? Visual Thinking in Unified Multimodal Models for Cross-View Spatial Reasoning},
  author={Yang, Qian and Sikarwar, Ankur and Le, Huy and Zhang, Le and Shi, Zhuan and Taslakian, Perouz and Agrawal, Aishwarya},
  journal={arXiv preprint arXiv:2605.27310},
  year={2026}
}

@article{guo2026toolbenefit,
  title={Do Multimodal Agents Really Benefit from Tool Use? A Systematic Study of Capability Gains},
  author={Guo, Garvin and Yu, Donglei and Chen, Yu and Wang, Xiang and Li, Shuai and Zhao, Xinpei and Liu, Huaxing and Wang, Qinghao and Liao, Minpeng},
  journal={arXiv preprint arXiv:2606.02357},
  year={2026}
}

@article{hu2026tvicot,
  title={{TVI-CoT}: Text-Visual Interleaved Chain-of-Thought Reasoning for Multimodal Understanding},
  author={Hu, Lianyu and Ma, Xiaoyu and Liao, Zeqin and Liu, Yang},
  journal={arXiv preprint arXiv:2606.08464},
  year={2026}
}

\clearpage
\beginappendix
\startcontents[app]
\begingroup
  \renewcommand{\contentsname}{Appendix Contents}
  \section*{\contentsname}
  \printcontents[app]{}{1}{}
\endgroup
\newpage
\section{Benchmark Construction Details}
\label{app:benchmark_details}

This section provides additional implementation details for constructing
See2ThinkBench. The benchmark construction consists of four stages:
source dataset collection, caption-only visual-dependency filtering,
answer-format conversion, and manual quality checking.

\subsection{Source Dataset Collection and Preprocessing}
\label{app:source_sampling}

We collect candidate samples from RBench-V~\citep{rbenchv},
EMMA~\citep{emma}, M3CoT~\citep{m3cot},
PRISM-Bench~\citep{prismbench}, CLEVR~\citep{johnson2017clevr},
SuperCLEVR~\citep{superclevr}, VLABench~\citep{vlabench},
DROID~\citep{droid}, and IntPhys2~\citep{intphys2}.

To enable unified evaluation, source-specific task names are mapped into the
semantic categories used in See2ThinkBench:

\begin{itemize}[leftmargin=*]
\item \textbf{2D structured reasoning:} Geometry, Spatial Puzzle, Physics,
Chemistry, Science QA, and Abstract Pattern, instantiated from RBench-V,
EMMA, M3CoT, and PRISM-Bench.
\item \textbf{3D scene reasoning:} Object Attributes and Compositional 3D,
instantiated from CLEVR and SuperCLEVR, respectively.
\item \textbf{Real-world visual reasoning:} Robot Manipulation, Robot State
Change, Visual Commonsense, and Intuitive Physics, instantiated from VLABench,
DROID, M3CoT, and IntPhys2, respectively.
\end{itemize}

We retain original images whenever available and modify only the textual
question and answer format. Before further processing, samples with missing
images, duplicated content, invalid references, corrupted files, or
insufficient visual evidence are removed.

\subsection{Caption-Only Visual-Dependency Filtering}
\label{app:visual_dependency_filtering}

We apply caption-only filtering to remove samples that can be solved without
direct access to the original image. For each candidate sample, a caption
generator receives both the image and the question, but is instructed to
describe only visually observable evidence without answering the question or
introducing unsupported inference.

A reasoning model then receives only the question and the generated caption,
while the original image is withheld. The model is queried five times. Samples
that are answered correctly in at least three trials are considered
caption-solvable and removed. Remaining samples are retained for subsequent
processing.

This procedure reduces text-dominant samples while preserving visually
dependent problems. It does not assume that all retained samples are
impossible to solve from language priors, but filters cases where textual
surrogates are sufficient.

\begin{promptbox}{Prompt for Detailed Image Caption Generation}

\textbf{System prompt:}

You are a precise multimodal description model. Describe all visually
observable information in the provided image that may be relevant to answering
questions about it.

Include visible objects, attributes, text, geometric structures, spatial
relations, diagram elements, and physical states. Do not answer the associated
question and do not infer information that is not visibly supported.

\textbf{User instruction prompt:}

Question associated with the image:

\{question\}

Generate a detailed and objective description of the image.

Return only the image description.

\textit{[The original image is provided.]}

\end{promptbox}

\begin{promptbox}{Prompt for Caption-Only Solvability Testing}

\textbf{System prompt:}

You are solving a visual-reasoning problem without access to the original
image. You are given only the question and a textual image description.
Use only the supplied information. Do not invent missing visual evidence.

\textbf{User instruction prompt:}

Question:

\{question\}

Image description:

\{caption\}

Reason carefully and return a concise final answer in the following format:

Final Answer: \{answer\}

\end{promptbox}

\subsection{Answer-Format Conversion}
\label{app:open_ended_conversion}

Many source datasets contain multiple-choice or template-based questions.
Whenever possible, we remove textual answer-option lists and rewrite the
question as a free-form query while preserving the original visual evidence and
reasoning requirement. A small subset (127 of 1,200 samples) retains structured
candidate sets inherited from source tasks because the alternatives themselves
form part of the reasoning problem; these candidates may appear in the question
text or as panels embedded in the original image. These instances are evaluated
as structured-answer tasks, while the remaining 1,073 use free-form answers.

For samples selected for free-form conversion, the rewritten question must
maintain a unique and reproducible reference answer. The conversion process
does not introduce additional evidence, reveal the correct answer, or change
the underlying task objective.

\begin{promptbox}{Prompt for Free-Form Question Rewriting}

\textbf{System prompt:}

You are converting a visual multiple-choice question into a free-form
question.

Preserve the original visual evidence and reasoning requirement. Remove the
answer choices and rewrite the question so that it has a unique and concise
answer.

Do not introduce new information. Do not reveal the correct option. Do not
change the underlying task.

\textbf{User instruction prompt:}

Original question:

\{original\_question\}

Answer options:

\{options\}

Correct answer:

\{correct\_answer\}

Return:

Open-ended question:
\{rewritten\_question\}

Reference answer:
\{reference\_answer\}

\textit{[The original image is provided.]}

\end{promptbox}

\subsection{Manual Quality Checking}
\label{app:manual_quality_control}

After automatic processing, retained samples undergo manual review.
Reviewers verify that the rewritten questions remain grounded in the images,
the reference answers are uniquely supported, and the visual evidence is
sufficient for answering the questions.

Samples are revised or removed if they contain ambiguous visual references,
underspecified answers, inconsistent image--question pairs, or substantial
text-only shortcuts.

Reviewers additionally verify that each sample naturally supports meaningful
visual operations, including highlighting relevant objects, cropping regions,
drawing auxiliary lines, or annotating visual relations.


\section{Inference Prompts and Implementation Details}
\label{app:vaot_prompts}

\subsection{Implementation Settings}
\label{app:implementation_settings}

We evaluate GPT-5.5, GPT-o3, Gemini 3.5 Flash, and
Qwen3-VL-32B-Instruct. For each model, all four inference settings use the
complete 1.2K-sample benchmark, the same original images, answer format, and
interaction budget. The main category-level outcome table and
Figure~\ref{fig:task_model_patterns} use all 1,200 samples, while group-level
aggregates are reported in Appendix~\ref{app:complete_outcome_results}.
Process scores use all 1.2K VAoT trajectories per model, for 4,800
trajectories in total. GPT-5.4 is used as the external answer and process
judge.

The renderer receives the current image state and the structured action only.
For faithful rendering, it is constrained to execute the requested operation
and preserve unrelated content. VAoT and VAoT-WrongRender use the same
reasoning prompt and interaction budget; WrongRender changes only the returned
visual state. In all main experiments, the intervention uses the
\texttt{modify\_key} condition to alter task-relevant evidence while aiming
to preserve a natural-looking appearance and the broad operation intent. The
reasoning model is not informed that the feedback has been corrupted. Paired
intervention analyses use complete sample-level alignments across all 1.2K
samples for all four models, as described in Section~\ref{sec:exp_setup}.

\subsection{Text-CoT Prompt}
\label{app:cot_prompt}
\begin{promptbox}{Prompt for Text-CoT}
\begin{lstlisting}[style=promptstyle]
ROLE AND GOAL
You are a multimodal reasoning assistant.
Your goal is to solve the provided problem by explicitly analyzing the given image(s) and generating complete reasoning steps with final answer.

INSTRUCTIONS
Observe carefully:
- Carefully inspect the provided image before reasoning.
- Describe what you observe that is relevant to the problem (e.g., spatial relations, geometric shapes, labels, or quantities).

Analyze the context:
- Review the original problem and integrate both visual and textual information.
- Generate a complete step-by-step reasoning process.

Generate complete solution:
- Provide all logical steps needed to solve the problem.
- Each step should build upon the previous one.
- End with the final answer.

Visual reasoning requirement:
- Your reasoning must explicitly reference visual evidence from the image(s).
- If the image were removed, your reasoning should be incomplete.
- Each step should reference what you see in the image that drives your reasoning.

OUTPUT FORMAT
Your output must contain the following parts:

**Complete Solution:**

**Step 1 (Text):**
Visual Observation: [Describe what the image shows relevant to this step]
Reasoning: [Concise explanation of your reasoning for this step]
Calculation/Result: [Show calculations and intermediate results]

**Step 2 (Text):**
Visual Observation: [Describe what the image shows relevant to this step]
Reasoning: [Concise explanation of your reasoning for this step]
Calculation/Result: [Show calculations and intermediate results]

[... continue for all steps ...]

**Final Answer:**
[The final numeric or symbolic result, clearly boxed or highlighted]

SPECIAL NOTES
- Generate all reasoning steps needed to solve the problem completely.
- Each step must explicitly reference visual elements from the image.
- Use mathematical notation where appropriate.
- The solution should be self-contained and logically complete.

EXAMPLE OUTPUT
**Complete Solution:**

**Step 1 (Text):**
Visual Observation: The image shows a right triangle with hypotenuse labeled as 10 cm and one acute angle labeled as 30 degrees.
Reasoning: In a right triangle, the side opposite the 30 degrees angle is half the length of the hypotenuse.
Calculation/Result: Opposite side = 10 cm  x  sin(30 degrees) = 10  x  0.5 = 5 cm

**Step 2 (Text):**
Visual Observation: The adjacent side to the 30 degrees angle can be found using the Pythagorean theorem or cosine function.
Reasoning: Using the cosine function: adjacent side = hypotenuse  x  cos(30 degrees)
Calculation/Result: Adjacent side = 10 cm  x  cos(30 degrees) = 10  x  (sqrt3/2) = 5sqrt3 cm ~= 8.66 cm

**Step 3 (Text):**
Visual Observation: The triangle has all three sides now determined.
Reasoning: The perimeter is the sum of all three sides.
Calculation/Result: Perimeter = 10 cm + 5 cm + 5sqrt3 cm = 15 + 5sqrt3 cm

**Final Answer:**
The perimeter of the triangle is 15 + 5sqrt3 cm

CONTEXT
{problem_statement}

YOUR TASK
Based on the context above, analyze the image(s) and generate a complete step-by-step solution with final answer, ensuring your reasoning depends explicitly on visual evidence.
\end{lstlisting}
\end{promptbox}

\subsection{VAoT Prompt}
\label{app:vaot_prompt}
\begin{promptbox}{Prompt for VAoT}
\begin{lstlisting}[style=promptstyle]
 **ROLE AND OBJECTIVE**  
You are an expert in **Multimodal Chain-of-Thought Reasoning**. Your goal is to solve problems step-by-step, acting as both a **logical reasoner** and a **visual designer**.

You have full autonomy to decide **when** a visual aid is *pedagogically essential* -- and **how expressively** to render it (e.g., callout labels, hand-drawn highlights), using publication-quality conventions.

---

**CORE PHILOSOPHY**

1. **Step-by-Step Logic**  
   Each step must represent **one atomic deduction**. Never collapse multiple ideas.

2. **Visual Necessity (CRITICAL)**  
   -> Generate a visual action **ONLY IF** it *uniquely enables understanding* that text alone cannot provide:  
   - [YES] **Highlighting a specific region** (e.g., `"stress concentration at joint"`)  
   - [YES] **Tracing an irregular path** (e.g., `"blood flow through capillary"`)  
   - [YES] **Zooming on unreadable details**  
   - [YES] **Showing spatial relationships**  
   - [YES] **Disambiguating overlapping objects**  
   -> **SKIP ACTION** for:  
   - Pure calculation (`"Sum forces: F1 + F2 = 10N"`)  
   - Logical inference (`"Since A > B and B > C, then A > C"`)  
   - Summary/conclusion without new visual reference  

3. **Clean & Expressive Communication**  
   - Labels **MUST** be semantic: `"Friction"`, not `"Box[200,300]"`  
   - When a visual *is* generated, **leverage enhanced styles** if they improve clarity:  
     - `callout` for key concepts  
     - `jitter` for organic structures  
     - Domain-appropriate `theme` (e.g., `biology` for cells)  
     -> But **never add decoration without purpose**.

**INSTRUCTIONS**

1. **Analyze Context**  
   Review `Problem`, `Original Image`, and `Previous Steps`.

2. **Identify Next Step**  
   What is the *single next* logical move?

3. **Determine Visual Necessity**

* **For Step 1**: Always output **Text Explanation + exactly one visual action object**.
  The `action` array MUST contain exactly one item.
  Do **not** output Final Answer in Step 1.

* **For Step 2 and later**:

  * Default to **Text Explanation only**.
  * Generate another visual action only when it helps inspect a new region, new object, new path, or new spatial relation that was not already shown.
  * Do not draw again for pure calculation, restatement, or conclusion.
  * Across the whole solution, prefer 1-3 visual actions unless the problem clearly needs more.


4. **Design Visual Action (If Chosen)**  
   Select the most effective tool -- and **optionally its expressive style**:

   | Use Case | Recommended Tool + Style |
   |---------|--------------------------|
   | Critical concept needing emphasis | `annotate` + `"shape": "callout"` + `"shadow": true` |
   | Irregular structure (vessel, river) | `trace_highlight` + `"jitter": true` + `"feather": "gradient"` |
   | Hypothetical boundary | `ellipse` + `"dash": "dashed"` |
   | Domain coherence | Add `global_style: { "theme": "biology" }` (once) |

**STRICT CONTENT RULES**

- **Coordinates**: Always normalized to **0-1000** scale.  
- **Labels**: **NO** coordinates, brackets, scores, or units.  
  - [NO] `"Force [300,400]"`, `"Mitochondrion (0.92)"`  
  - [YES] `"Force"`, `"Mitochondrion"`  
- **Safety**: If `style`/`shape` fields are invalid -> renderer ignores them silently.

**AVAILABLE VISUAL ACTIONS**

| Action | Required | Optional Fields | Notes |
|-------|----------|----------------|-------|
| `annotate` | `target`, `content` | `shape`, `style` | `shape`: `"box"` (default), `"ellipse"`, `"callout"` |
| `trace_highlight` | `target`, `content` | `style` | `target`: 3-5 spine points; `style`: `{jitter, feather, opacity}` |
| `highlight` / `mask` | `target` | `style` | `style`: `{feather, texture}` |
| `overlay_text` | `target`, `content` | `style` | `style`: `{font_weight, bg_adaptive}` |
| `draw_line` | `target`, `content` | `style` | -- |
| `ellipse` | `target`, `content` | `style` | -- |
| `crop` / `zoom` | `target` | -- | -- |

**Global Style (Optional, Top-Level)**  
Add once (e.g., Step 1) to set rendering context:
```json
"global_style": { "theme": "biology", "enable_decor": true }
```

**OUTPUT FORMAT (Strict)**

**Step N (Text):**  
[A concise reasoning. Explain WHAT, WHY, and *whether visual is necessary*. Optionally suggest style/theme.]

> *Example:*  
> *"The capillary network is irregular and directional; a bounding box would misrepresent its flow. A trace highlight with gradient fade better conveys blood movement. Visual is necessary to show path topology."*

**Step N (Action Description):**  
*(INCLUDE ONLY IF visual is deemed necessary)*

```json
{
  "step": N,
  "visual_rationale": "Why this tool/style? (e.g., 'callout for emphasis on key term')",
  "global_style": { "theme": "biology" },
  "action": [
    {
      "type": "trace_highlight",
      "target": [[320,210], [380,240], [450,260]],
      "content": "Capillary Flow",
      "style": { "jitter": true, "feather": "gradient" }
    },
    {
      "type": "annotate",
      "shape": "callout",
      "target": [300,200,340,240],
      "content": "O2 Exchange",
      "style": { "corner_radius": 6 }
    }
  ]
}
```

**RULES FOR TERMINATION**

- Continue until solution is complete.  
- Final step:  
  **Final Answer:** [The Answer]

**CONTEXT**  
Problem:  
{problem_statement}  

Previous Steps:  
{previous_steps}  

**YOUR TASK**  
Based on the context above, generate the single next step. Decide strictly whether a visual action is necessary for this step.
\end{lstlisting}
\end{promptbox}

\subsection{VAoT-NoRender Prompt}
\label{app:norender_prompt}
\begin{promptbox}{Prompt for VAoT-NoRender}
\begin{lstlisting}[style=promptstyle]
**ROLE AND OBJECTIVE**
You are an expert in Multimodal Chain-of-Thought Reasoning.

This is the VAoT-NoRender setting. You may propose visual actions, but the system will not render them. Therefore, proposed actions are only action-text suggestions and must not be treated as new visual evidence.

**CORE RULES**

1. Generate only the single next step.
2. Each step should make one atomic reasoning move.
3. Reason only from the problem statement, the original image, and previous text steps.
4. You may propose a visual action when it would help, but do not claim it has been executed.
5. Do not say that a region is highlighted, zoomed, cropped, traced, or annotated unless that state was already present in the original image.
6. Do not use previous proposed actions as evidence.
7. For embodied control tasks, predict only the next immediate requested action.

**WHEN TO PROPOSE A VISUAL ACTION**
Propose a visual action only when it would clearly help locate a key region, trace a path, show a spatial relation, separate overlapping objects, or zoom into dense details.

Skip visual actions for pure calculation, logical inference, summary, or final answer.

**SUPPORTED VISUAL ACTIONS**
Use only these action types:
`annotate`, `highlight`, `mask`, `trace_highlight`, `draw_line`, `ellipse`, `crop`, `zoom`, `overlay_text`.

Coordinates must be normalized to 0-1000. Labels must be semantic and human-readable.

**OUTPUT FORMAT**

**Step N (Text):**
[One concise reasoning step. State what is observed, why it matters, and whether a visual action would be useful.]

**Step N (Action Description):**
Include this block only if a visual action would be useful.

```json
{
  "step": N,
  "render_policy": "no_render",
  "visual_rationale": "Why this proposed action would help, although it is not rendered.",
  "action": [
    {
      "type": "annotate",
      "target": [300, 200, 340, 240],
      "content": "Key Evidence"
    }
  ]
}
```

When ready, end with:

**Final Answer:** [The answer]

**CONTEXT**
Problem:
{problem_statement}

Previous Steps:
{previous_steps}

**YOUR TASK**
Generate the single next step. Remember: proposed actions are not rendered and must not be used as visual evidence.
\end{lstlisting}
\end{promptbox}

\subsection{WrongRender Intervention Prompt}
\label{app:wrongrender_prompt}

VAoT-WrongRender uses the same reasoning prompt and interaction procedure as
standard VAoT. The intervention is introduced only at the rendering stage:
after the reasoning model proposes a visual action, the image editor receives
the following prompt to generate a natural-looking but intentionally corrupted
visual state. The main evaluation uses only the \texttt{modify\_key} condition,
which alters task-relevant visual evidence. The supported
\texttt{modify\_non\_key} condition is not used in the experiments reported in
this paper.

\begin{promptbox}{Prompt for WrongRender Intervention}
You are an image interference expert. Your task is to modify the provided image
according to the interference type.

Original problem: \{problem\}

Interference type: \{interference\_type\}

Please modify this image according to the interference type:
\begin{enumerate}[leftmargin=*]
\item If the interference type is \texttt{modify\_key}, please modify key
areas in the image so that it produces errors or misleading information in
important aspects that affect problem solving.
\item If the interference type is \texttt{modify\_non\_key}, please modify
background or non-key information in the image, keeping the important
information correct but creating interference in secondary details.
\end{enumerate}

Please analyze the image content, identify the key versus non-key elements for
solving the problem, and apply appropriate modifications based on the
interference type. The modified image should look natural but contain the
intended interference.
\end{promptbox}


\clearpage

\section{Detailed Evaluation Protocols}
\label{app:evaluation_protocols}

\subsection{Final-Answer Evaluation}
\label{app:answer_evaluation}

We first extract the final answer. Structured-answer outputs are evaluated by
exact matching after deterministic normalization of case, whitespace,
punctuation, unit formatting, and common numerical forms. Free-form predictions
that require semantic comparison are judged against the reference answer with
the following production prompt.

\begin{promptbox}{Prompt for Accuracy / Answer Evaluation}
\begin{lstlisting}[style=promptstyle,language=Python]
You are a strict answer evaluator for a multimodal reasoning benchmark. Judge whether the model's final answer is semantically correct with respect to the reference answer.

Question:
{row.get("question", "")}

Reference answer:
{row.get("ground_truth", "")}

Model final answer:
{row.get("final_answer", "")}

Rules:
1. Ignore harmless formatting differences, units formatting, LaTeX wrappers, punctuation, and equivalent wording.
2. For math, accept algebraically equivalent expressions and numerically equivalent answers when the intended quantity matches.
3. For multiple-choice questions, accept either the correct option letter or the correct option text.
4. If the model gives extra claims that contradict the reference, mark it incorrect.
5. If the model answer is empty, evasive, or says it cannot answer, mark it incorrect.
6. Be strict about counts, directions, relations, and named entities.

Return only a JSON object with:
{{
  "correct": true or false,
  "reason": "brief explanation"
}}
\end{lstlisting}
\end{promptbox}

\subsection{Process-Level Evaluation}
\label{app:process_judge}

For every VAoT trajectory, the judge receives the question, reference
answer, model final answer, and complete trajectory. The final response is
provided because Feedback Uptake requires assessing whether information from
the rendered visual state is incorporated into subsequent reasoning and the
model's final response. The judge selects one key visual step and scores Action
Relevance, Render Faithfulness, and Feedback Uptake on $\{0,0.5,1\}$.
Model identity is withheld, and no explicit correct/incorrect label is
supplied.

\begin{promptbox}{Prompt for Key-Step Selection and Process Evaluation}
\begin{lstlisting}[style=promptstyle]
You are an expert evaluator for multimodal reasoning trajectories.

You will evaluate a Visual Action-of-Thought (VAoT) trajectory. The model solves a visual reasoning problem by alternating between textual thoughts, visual actions, rendered visual states, and subsequent reasoning.

Your task is to evaluate ONLY the VAoT trajectory. Do NOT compare it with Text-CoT, VAoT-NoRender, or VAoT-WrongRender.

Question:
{question}

Ground-truth answer:
{ground_truth}

Model final answer:
{final_answer}

Trajectory:
{trajectory_text}

Please first select the key visual step, then evaluate that selected step along three dimensions.

1. Key Visual Step Selection:
Identify the single visual step that is most relevant to the final reasoning or most directly exposes the failure source.
If there is no effective visual operation, use null.

1. Action Relevance:
Does the model choose visual actions that are useful for solving the task?
Score 0 / 0.5 / 1.
- 1: The visual action directly targets task-relevant evidence and is useful for solving the question.
- 0.5: The visual action is partially relevant, weakly targeted, or contains unnecessary but not harmful operations.
- 0: The visual action is irrelevant, generic, decorative, or does not meaningfully support the task.

1. Render Faithfulness:
Are the rendered visual states faithful to the intended visual actions?
Score 0 / 0.5 / 1.
- 1: The rendered visual state faithfully executes the intended action.
- 0.5: The rendering partially matches the action, but has noticeable ambiguity, imprecision, or minor errors.
- 0: The rendering is missing, wrong, misleading, or inconsistent with the intended action.

1. Feedback Uptake:
Does the subsequent reasoning actually use the rendered visual states?
Score 0 / 0.5 / 1.
- 1: Subsequent reasoning clearly uses the rendered visual state as evidence.
- 0.5: Subsequent reasoning weakly or implicitly uses the rendered state, but still relies mostly on text priors.
- 0: Subsequent reasoning ignores the rendered state or continues independently of it.

Return ONLY valid JSON in the following format:

{{
  "key_step_id": <step id or null>,
  "key_step_reason": "<short explanation>",
  "action_relevance": {{
    "score": <0, 0.5, or 1>,
    "reason": "<short explanation>"
  }},
  "render_faithfulness": {{
    "score": <0, 0.5, or 1>,
    "reason": "<short explanation>"
  }},
  "feedback_uptake": {{
    "score": <0, 0.5, or 1>,
    "reason": "<short explanation>"
  }},
  "overall_failure_source": "<action_relevance | render_faithfulness | feedback_uptake | none | unclear>",
  "summary": "<one-sentence diagnostic summary>"
}}
\end{lstlisting}
\end{promptbox}

\subsection{Paired Intervention Analysis}
\label{app:statistical_analysis}

We perform paired sample-level analysis for two comparisons:
\begin{itemize}[leftmargin=*]
\item VAoT-NoRender $\rightarrow$ VAoT, which separates rendering benefit
and rendering harm;
\item VAoT $\rightarrow$ VAoT-WrongRender, which measures accuracy
degradation under corrupted visual feedback.
\end{itemize}
For the first comparison, transition shares are normalized by all matched
trajectories within each Render Faithfulness group. For the second,
trajectories are grouped by Feedback Uptake and we report
$\mathrm{Acc}_{\mathrm{VAoT}}-\mathrm{Acc}_{\mathrm{WrongRender}}$.

\FloatBarrier

\clearpage

\section{Complete Outcome and Paired-Intervention Results}
\label{app:complete_outcome_results}

\subsection{Complete 1.2K Outcome Results}
Table~\ref{tab:full_outcome_results} reports group-level aggregates for all four
settings on the complete benchmark. The main paper reports the corresponding
12-category results on the same 1,200 samples; this table provides a compact
summary by visual-reasoning environment.

\subsection{Complete Paired-Intervention Statistics}
Tables~\ref{tab:rf_transition_full} and~\ref{tab:fu_sensitivity_full} provide
the counts and common-denominator rates underlying
Figures~\ref{fig:render_transition} and~\ref{fig:uptake_corruption}.
Table~\ref{tab:semantic_answer_change} reports semantic answer changes between
VAoT and VAoT-WrongRender. These paired statistics use all four evaluated
models, for which complete sample-level alignment records are retained.

\begin{table}[H]
\centering
\small
\setlength{\tabcolsep}{9pt}
\renewcommand{\arraystretch}{1.06}
\caption{\textbf{Accuracy on the complete 1.2K-sample benchmark.}}
\label{tab:full_outcome_results}
\begin{tabular*}{\textwidth}{@{\extracolsep{\fill}}llrrrr@{}}
\toprule
\textbf{Model} & \textbf{Setting} & \textbf{2D} & \textbf{3D}
& \textbf{Real-world} & \textbf{Overall} \\
\midrule
GPT-5.5 & CoT & 55.33 & 52.50 & 41.25 & 50.17 \\
 & VAoT-NoRender & 50.00 & 57.00 & 40.50 & 48.00 \\
 & VAoT & 45.33 & 52.50 & 39.50 & 44.58 \\
 & VAoT-WrongRender & 43.83 & 47.50 & 37.25 & 42.25 \\
GPT-o3 & CoT & 40.50 & 63.50 & 38.00 & 43.50 \\
 & VAoT-NoRender & 36.50 & 64.00 & 35.75 & 40.83 \\
 & VAoT & 36.67 & 67.50 & 42.50 & 43.75 \\
 & VAoT-WrongRender & 36.83 & 48.00 & 38.75 & 39.33 \\
Gemini 3.5 Flash & CoT & 61.83 & 77.50 & 39.50 & 57.00 \\
 & VAoT-NoRender & 61.33 & 80.50 & 42.25 & 58.17 \\
 & VAoT & 60.33 & 77.50 & 39.25 & 56.17 \\
 & VAoT-WrongRender & 59.17 & 64.50 & 40.75 & 53.92 \\
Qwen3-VL-32B-Instruct & CoT & 32.17 & 60.00 & 37.75 & 38.67 \\
 & VAoT-NoRender & 30.33 & 58.00 & 37.00 & 37.17 \\
 & VAoT & 29.67 & 57.00 & 34.00 & 35.67 \\
 & VAoT-WrongRender & 31.67 & 40.00 & 36.75 & 34.75 \\
\bottomrule
\end{tabular*}
\end{table}

\begin{table}[H]
\centering
\small
\setlength{\tabcolsep}{8pt}
\renewcommand{\arraystretch}{1.05}
\caption{\textbf{Rendering benefit and harm on the four-model aligned subset.}
Benefit and harm are percentages of all aligned trajectories within each task
group and Render Faithfulness bin; Net is Benefit minus Harm.}
\label{tab:rf_transition_full}
\begin{tabular*}{\textwidth}{@{\extracolsep{\fill}}lrrrrr@{}}
\toprule
\textbf{Task group} & \textbf{RF} & \textbf{$n$} & \textbf{Benefit (\%)}
& \textbf{Harm (\%)} & \textbf{Net (pp)} \\
\midrule
2D Structured & 0 & 99 & 10.1 & 16.2 & -6.1 \\
 & 0.5 & 1788 & 9.6 & 11.0 & -1.4 \\
 & 1 & 513 & 9.4 & 10.5 & -1.2 \\
3D Scene & 0 & 140 & 6.4 & 15.7 & -9.3 \\
 & 0.5 & 344 & 12.8 & 12.2 & +0.6 \\
 & 1 & 316 & 8.5 & 8.2 & +0.3 \\
Real-world & 0 & 46 & 8.7 & 6.5 & +2.2 \\
 & 0.5 & 1008 & 8.7 & 9.4 & -0.7 \\
 & 1 & 546 & 8.4 & 7.5 & +0.9 \\
\bottomrule
\end{tabular*}
\end{table}

\begin{table}[H]
\centering
\small
\setlength{\tabcolsep}{7pt}
\renewcommand{\arraystretch}{1.05}
\caption{\textbf{Sensitivity to corrupted feedback on the four-model aligned subset.}
Accuracy Drop is $\mathrm{Acc}_{\mathrm{VAoT}}-\mathrm{Acc}_{\mathrm{WrongRender}}$.}
\label{tab:fu_sensitivity_full}
\begin{tabular*}{\textwidth}{@{\extracolsep{\fill}}lrrrrr@{}}
\toprule
\textbf{Task group} & \textbf{FU} & \textbf{$n$}
& \textbf{VAoT Acc. (\%)} & \textbf{Wrong Acc. (\%)}
& \textbf{Drop (pp)} \\
\midrule
Overall & 0 & 607 & 42.2 & 44.5 & -2.3 \\
 & 0.5 & 786 & 42.9 & 39.4 & +3.4 \\
 & 1 & 3407 & 46.1 & 42.9 & +3.1 \\
2D Structured & 0 & 385 & 41.6 & 46.0 & -4.4 \\
 & 0.5 & 403 & 44.2 & 43.4 & +0.7 \\
 & 1 & 1612 & 43.1 & 42.0 & +1.1 \\
3D Scene & 0 & 86 & 54.7 & 51.2 & +3.5 \\
 & 0.5 & 87 & 60.9 & 50.6 & +10.3 \\
 & 1 & 627 & 65.2 & 49.8 & +15.5 \\
Real-world & 0 & 136 & 36.0 & 36.0 & -0.0 \\
 & 0.5 & 296 & 35.8 & 30.7 & +5.1 \\
 & 1 & 1168 & 39.9 & 40.6 & -0.7 \\
\bottomrule
\end{tabular*}
\end{table}

\begin{table}[H]
\centering
\small
\setlength{\tabcolsep}{12pt}
\renewcommand{\arraystretch}{1.05}
\caption{\textbf{Semantic answer changes under corrupted feedback.}
Changed answers are determined by the semantic answer-change judge rather
than direct string comparison.}
\label{tab:semantic_answer_change}
\begin{tabular}{lrr}
\toprule
\textbf{Model} & \textbf{Changed} & \textbf{Change Rate (\%)} \\
\midrule
GPT-5.5 & 624 / 1,200 & 52.0 \\
GPT-o3 & 607 / 1,200 & 50.6 \\
Gemini 3.5 Flash & 392 / 1,200 & 32.7 \\
Qwen3-VL-32B-Instruct & 650 / 1,200 & 54.2 \\
\bottomrule
\end{tabular}
\end{table}
\FloatBarrier


\clearpage

\section{Complete Process Statistics}
\label{app:complete_process_statistics}

Table~\ref{tab:process_by_correctness} reports the process scores used to
compute the correct-minus-incorrect gaps in
Figure~\ref{fig:process_score_patterns}(b). Counts correspond to VAoT
final-answer outcomes for all four evaluated models on the complete
benchmark.

\begin{table}[H]
\centering
\scriptsize
\setlength{\tabcolsep}{6pt}
\renewcommand{\arraystretch}{1.05}
\caption{\textbf{Process scores split by final-answer correctness.}
Action, Render, and Feedback denote Action Relevance, Render Faithfulness, and
Feedback Uptake, respectively. Results use the complete four-model benchmark.}
\label{tab:process_by_correctness}
\begin{tabular*}{\linewidth}{@{\extracolsep{\fill}}lllrrrr@{}}
\toprule
\textbf{Model} & \textbf{Answer} & \textbf{Group} & \textbf{Count}
& \textbf{Action} & \textbf{Render} & \textbf{Feedback} \\
\midrule
\multirow{6}{*}{GPT-5.5}
& \multirow{3}{*}{Correct}   & 2D         & 272 & 0.9798 & 0.6085 & 0.7610 \\
&                            & 3D         & 105 & 0.9952 & 0.5905 & 0.7905 \\
&                            & Real-world & 158 & 0.9937 & 0.6646 & 0.8734 \\
& \multirow{3}{*}{Incorrect} & 2D         & 328 & 0.9802 & 0.5838 & 0.6921 \\
&                            & 3D         &  95 & 0.9947 & 0.5053 & 0.7789 \\
&                            & Real-world & 242 & 0.9835 & 0.6921 & 0.8140 \\
\midrule
\multirow{6}{*}{GPT-o3}
& \multirow{3}{*}{Correct}   & 2D         & 220 & 0.9909 & 0.6091 & 0.8318 \\
&                            & 3D         & 135 & 0.9963 & 0.6519 & 0.8815 \\
&                            & Real-world & 170 & 1.0000 & 0.6206 & 0.8529 \\
& \multirow{3}{*}{Incorrect} & 2D         & 380 & 0.9961 & 0.5763 & 0.8000 \\
&                            & 3D         &  65 & 0.9846 & 0.4692 & 0.7615 \\
&                            & Real-world & 230 & 0.9087 & 0.6957 & 0.8522 \\
\midrule
\multirow{6}{*}{Gemini 3.5 Flash}
& \multirow{3}{*}{Correct}   & 2D         & 362 & 0.9820 & 0.5994 & 0.6657 \\
&                            & 3D         & 155 & 0.9871 & 0.7097 & 0.8161 \\
&                            & Real-world & 157 & 0.9904 & 0.5987 & 0.7389 \\
& \multirow{3}{*}{Incorrect} & 2D         & 238 & 0.9622 & 0.5987 & 0.6408 \\
&                            & 3D         &  45 & 0.9889 & 0.6889 & 0.8000 \\
&                            & Real-world & 243 & 0.9609 & 0.6646 & 0.6605 \\
\midrule
\multirow{6}{*}{Qwen3-VL-32B-Instruct}
& \multirow{3}{*}{Correct}   & 2D         & 178 & 0.9522 & 0.5843 & 0.8539 \\
&                            & 3D         & 114 & 0.9912 & 0.6009 & 0.9386 \\
&                            & Real-world & 136 & 0.9779 & 0.5772 & 0.8824 \\
& \multirow{3}{*}{Incorrect} & 2D         & 422 & 0.9609 & 0.5533 & 0.8223 \\
&                            & 3D         &  86 & 0.9593 & 0.5814 & 0.8779 \\
&                            & Real-world & 264 & 0.9337 & 0.6742 & 0.9223 \\
\bottomrule
\end{tabular*}
\end{table}
\FloatBarrier


\section{Human Validation and WrongRender Quality Audit}
\label{app:human_validation}

\subsection{Human Validation of Process Annotations}
\label{app:human_process_validation}

The human audit is a diagnostic reliability check rather than a second full
annotation benchmark. Two annotators independently review the same 240
VAoT trajectories, sampled with a $4\times3\times4\times5$ design over
GPT-5.5, GPT-o3, Gemini 3.5 Flash, and Qwen3-VL-32B-Instruct, three task
groups, four diagnostic strata, and five trajectories per cell. The four strata are:
\begin{enumerate}[leftmargin=*]
\item Action Relevance $=1$ with an incorrect VAoT answer;
\item Action Relevance $=1$ and Render Faithfulness $=0$;
\item Action Relevance $=1$ and Render Faithfulness $=0.5$;
\item Action Relevance in $\{0,0.5\}$.
\end{enumerate}
This construction covers high-action/failed-answer cases, rendering failures,
partial renders, and weak action selection across all audited models and environments.

For each trajectory, the annotators inspect the original image and question,
the complete VAoT trajectory, the judge-selected key step, the three
process scores, and the judge explanations. Each target is labeled
\textit{reasonable}, \textit{partially reasonable}, or
\textit{unreasonable}. The main paper reports rates pooled across both annotators; the
per-annotator summaries are shown in Table~\ref{tab:human_by_annotator}.

\begin{table*}[t]
\centering
\small
\caption{\textbf{Human-validation results by annotator.}
R denotes Reasonable; R/P denotes Reasonable or Partially Reasonable. Each
annotator independently reviews the same 240 trajectories.}
\label{tab:human_by_annotator}
\begin{tabular*}{\textwidth}{@{\extracolsep{\fill}}lrrrr@{}}
\toprule
\textbf{Annotation target} & \textbf{A1 R} & \textbf{A1 R/P}
& \textbf{A2 R} & \textbf{A2 R/P} \\
\midrule
Key-step selection  & 88.8\% & 94.2\% & 87.9\% & 94.6\% \\
Action relevance    & 87.1\% & 96.7\% & 62.5\% & 97.1\% \\
Render faithfulness & 83.8\% & 93.3\% & 56.7\% & 92.5\% \\
Feedback uptake     & 87.5\% & 95.4\% & 71.7\% & 94.6\% \\
\bottomrule
\end{tabular*}
\end{table*}
\FloatBarrier

\begin{promptbox}{Human Review Instructions}
You are reviewing an automatic evaluation of a visual-reasoning trajectory.
For key-step selection, determine whether the selected step captures the visual
operation most relevant to the final reasoning. For each process dimension,
assess whether the score and explanation are supported by the complete
trajectory.

\textbf{Reasonable:} the judgment is clearly supported.\par
\textbf{Partially reasonable:} the judgment captures the main behavior but is
incomplete, imprecise, or slightly over- or under-scored.\par
\textbf{Unreasonable:} the judgment is contradicted by the trajectory or misses
the central behavior.

For Feedback Uptake, credit only information that is actually present in the
rendered state; do not infer uptake merely because the requested action is
relevant.
\end{promptbox}

The annotators differ most at the strict Reasonable-versus-Partially-Reasonable
boundary for Action Relevance and Render Faithfulness. For every target, both
annotators nevertheless mark more than 92\% of judgments as at least partially
reasonable.

\subsection{WrongRender Quality Audit}
\label{app:wrongrender_audit}

We manually audit 120 VAoT-WrongRender cases. The original audit covers 90
cases sampled uniformly across GPT-5.5, GPT-o3, and Gemini 3.5 Flash and the
three task groups, with 10 cases per model--group cell; we additionally audit
30 Qwen3-VL-32B-Instruct cases using the same task-group coverage. The audit
examines whether a corrupted state changes task-relevant content in an
incorrect direction while remaining plausible as an output from the same
renderer. Each case is assessed along five criteria: corruption validity,
visual plausibility, operation consistency, task relevance, and preservation
of unrelated image content.

\begin{promptbox}{Human Audit Instructions for WrongRender}
Inspect the original image, requested visual action, correct rendered state,
and WrongRender state. For each criterion below, assign Pass, Partial, or Fail
and briefly explain failed criteria.
\begin{itemize}[leftmargin=*]
\item \textbf{Corruption validity:} task-relevant visual content is changed in
an incorrect direction.
\item \textbf{Plausibility:} the result resembles a plausible output from the
same renderer.
\item \textbf{Operation consistency:} the requested operation type is
preserved.
\item \textbf{Task relevance:} the corruption could affect reasoning for the
question.
\item \textbf{Content preservation:} unrelated image regions remain
substantially unchanged.
\end{itemize}
\end{promptbox}

A case receives the aggregate label \textbf{Strict Pass} when none of the five
criteria is labeled Fail. It is considered \textbf{Acceptable} when at most
one criterion is labeled Fail. Table~\ref{tab:wrongrender_audit_overall}
reports the aggregate quality rates before and after adding the 30 Qwen cases.
The combined 120-case audit yields 56.7\% Strict Pass and 78.3\% Acceptable
rates. The added Qwen cases are somewhat more reliable than the original
three-model audit, with 76.7\% Strict Pass and 86.7\% Acceptable rates. We
therefore treat WrongRender as an aggregate diagnostic intervention rather
than assuming that every corrupted state realizes the intended modification
equally well.

\begin{table}[H]
\centering
\small
\setlength{\tabcolsep}{10pt}
\renewcommand{\arraystretch}{1.08}
\caption{\textbf{WrongRender quality-audit summary after adding Qwen.}}
\label{tab:wrongrender_audit_overall}
\begin{tabular}{lccc}
\toprule
\textbf{Subset} & \textbf{$N$} & \textbf{Pass (0 Fail)} & \textbf{Pass-or-Partial ($\leq$1 Fail)} \\
\midrule
Original audit & 90 & 50.0\% (45/90) & 75.6\% (68/90) \\
Qwen addition & 30 & 76.7\% (23/30) & 86.7\% (26/30) \\
Combined & 120 & 56.7\% (68/120) & 78.3\% (94/120) \\
\bottomrule
\end{tabular}
\end{table}

\paragraph{Paired analysis on quality-filtered subsets.}
To examine whether the observed VAoT--WrongRender accuracy gap is driven
primarily by failed corruptions, we recompute the paired accuracy difference
on the combined 120-case human-audited set. The strict subset contains only
cases with no failed audit criteria, while the relaxed subset allows at most
one failed criterion. VAoT and VAoT-WrongRender are compared on the same
samples within each subset.

\begin{table}[H]
\centering
\small
\setlength{\tabcolsep}{8pt}
\renewcommand{\arraystretch}{1.08}
\caption{\textbf{Paired accuracy on human-audited WrongRender subsets.}
Drop denotes
$\mathrm{Acc}_{\mathrm{VAoT}}-\mathrm{Acc}_{\mathrm{WrongRender}}$
in percentage points.}
\label{tab:wrongrender_filtered_accuracy}
\begin{tabular}{lcccc}
\toprule
\textbf{Subset}
& \textbf{$N$}
& \textbf{VAoT Acc.}
& \textbf{WrongRender Acc.}
& \textbf{Drop} \\
\midrule
Strict Pass (0 Fail)
& 68
& 42.65
& 33.82
& 8.82 \\

Acceptable ($\leq 1$ Fail)
& 94
& 41.49
& 38.30
& 3.19 \\
\bottomrule
\end{tabular}
\end{table}

The Qwen-only strict and acceptable subsets show accuracy drops of 8.70 and
7.69 percentage points, respectively. After combining all four models, the
strict subset exhibits an 8.82-point drop, while the relaxed subset retains a
positive 3.19-point drop. The behavioral-dependence result is therefore not
restricted to interventions that fail the quality audit, but its estimated
magnitude remains sensitive to intervention quality.

\FloatBarrier


\clearpage

\section{Qualitative Trajectories}
\label{app:qualitative_cases}

We provide representative trajectories spanning 2D structured, 3D scene, and
real-world visual reasoning. The examples are selected to illustrate three
complementary behaviors: successful visual-state use, process-level failures,
and answer changes under corrupted feedback. All panels use the original
model-produced images without redrawing.

\subsection{Successful Visual-State Use}
\label{app:qualitative_success}

\begin{figure}[H]
\centering
\textbf{2D structured reasoning: resistor-network comparison (GPT-5.5)}\par\vspace{2mm}
\begin{minipage}[t]{0.48\textwidth}
  \centering
  \includegraphics[width=\linewidth,height=3.0cm,keepaspectratio]{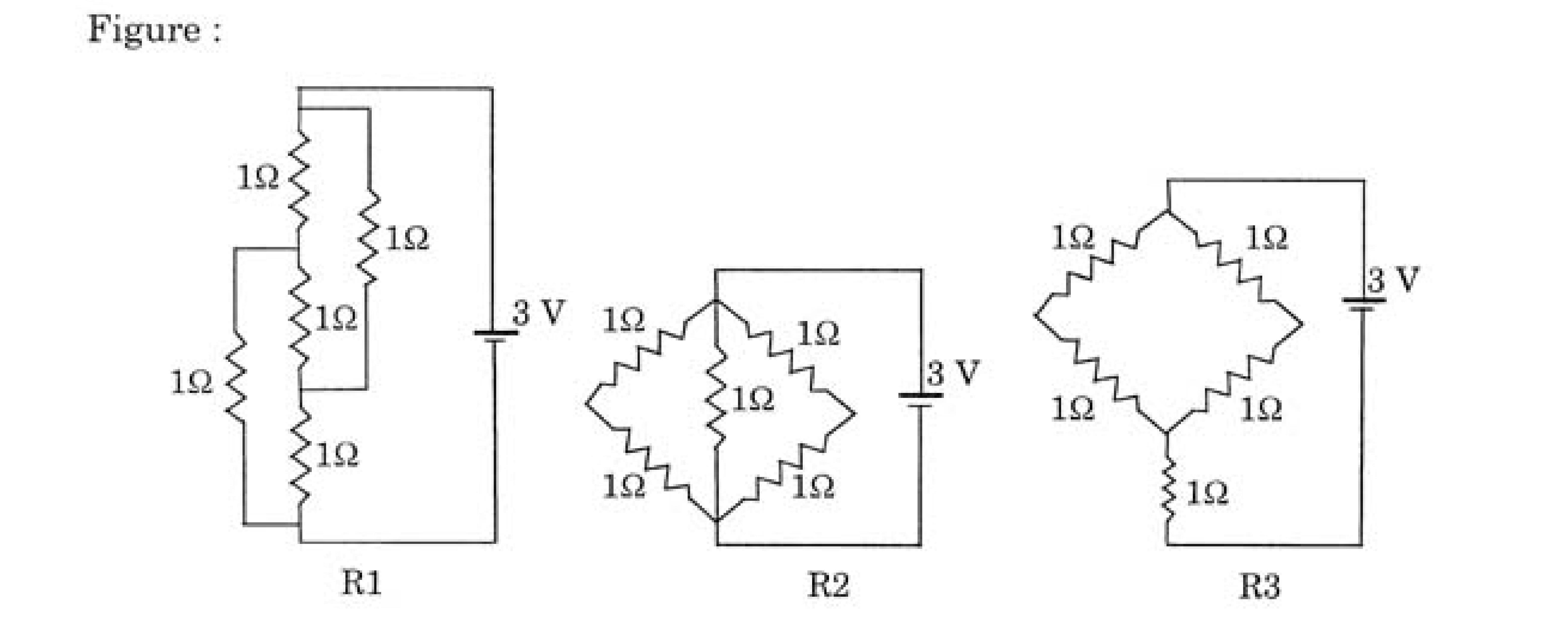}\\[0.6mm]
  {\small Original image}
\end{minipage}
\hfill
\begin{minipage}[t]{0.48\textwidth}
  \centering
  \includegraphics[width=\linewidth,height=3.0cm,keepaspectratio]{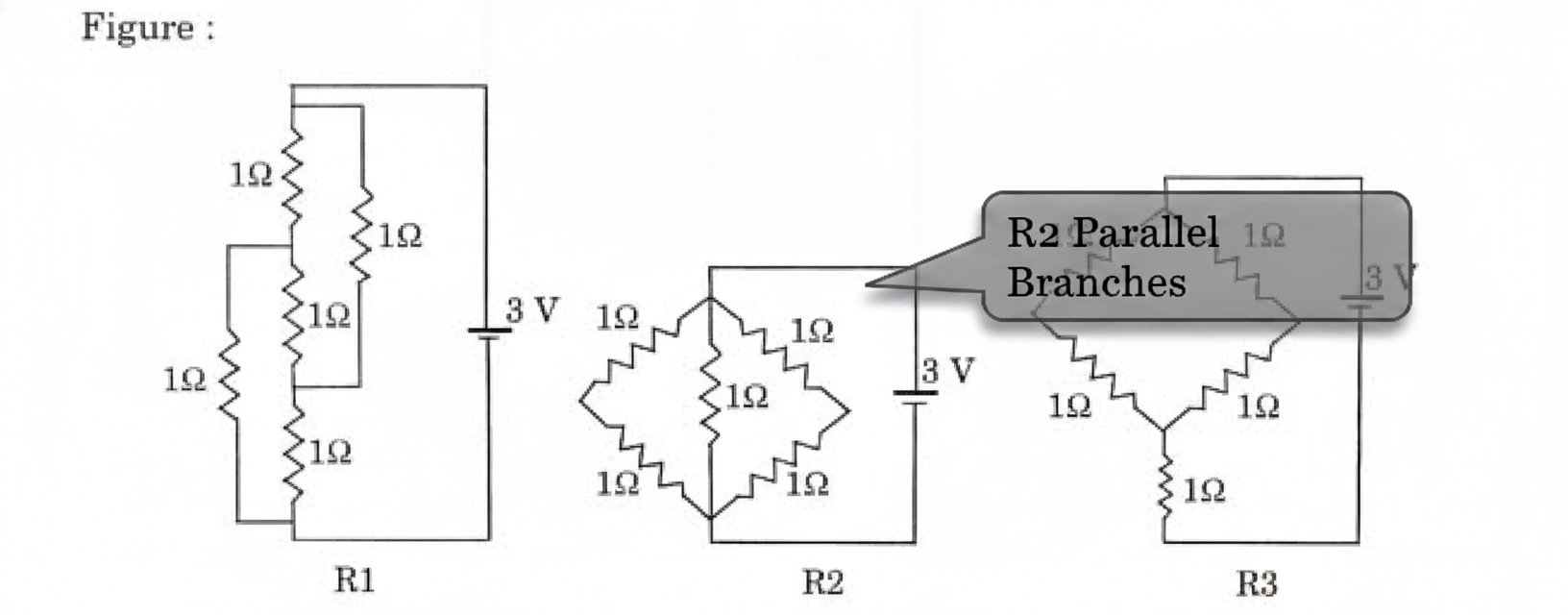}\\[0.6mm]
  {\small Rendered visual state}
\end{minipage}

\vspace{4mm}
\textbf{3D scene reasoning: relation-based object counting (GPT-o3)}\par\vspace{2mm}
\begin{minipage}[t]{0.48\textwidth}
  \centering
  \includegraphics[width=\linewidth,height=3.2cm,keepaspectratio]{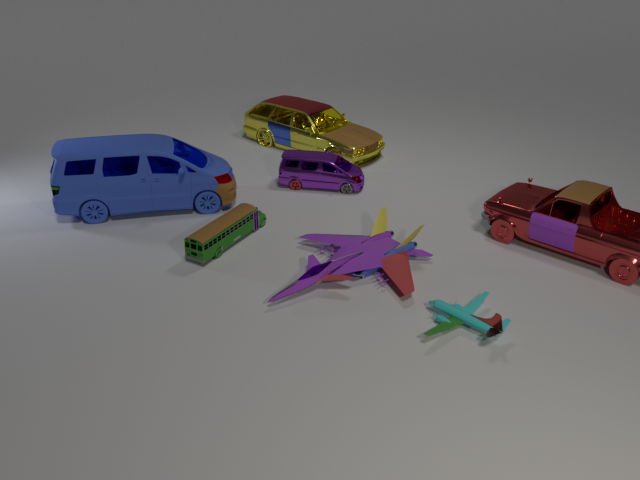}\\[0.6mm]
  {\small Original image}
\end{minipage}
\hfill
\begin{minipage}[t]{0.48\textwidth}
  \centering
  \includegraphics[width=\linewidth,height=3.2cm,keepaspectratio]{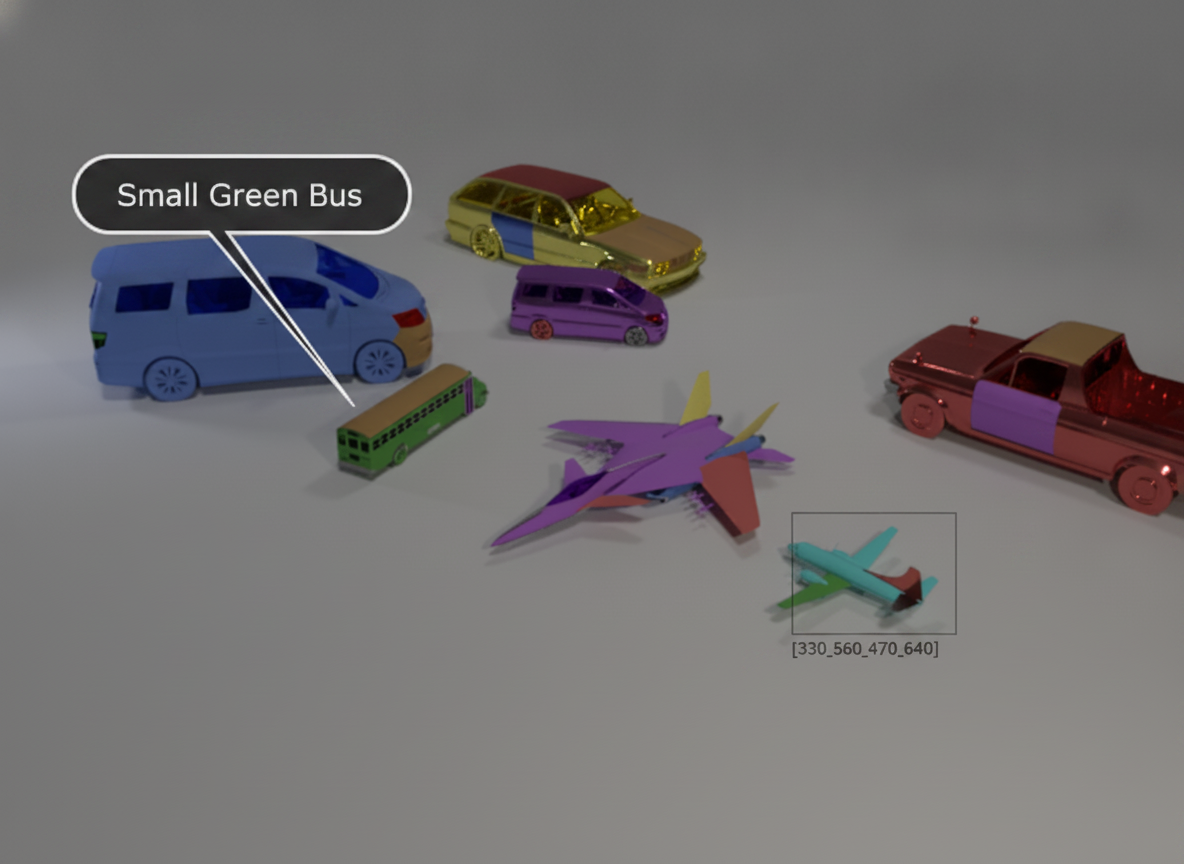}\\[0.6mm]
  {\small Rendered visual state}
\end{minipage}

\vspace{4mm}
\textbf{Real-world reasoning: activity recognition (Gemini 3.5 Flash)}\par\vspace{2mm}
\begin{minipage}[t]{0.48\textwidth}
  \centering
  \includegraphics[width=\linewidth,height=3.2cm,keepaspectratio]{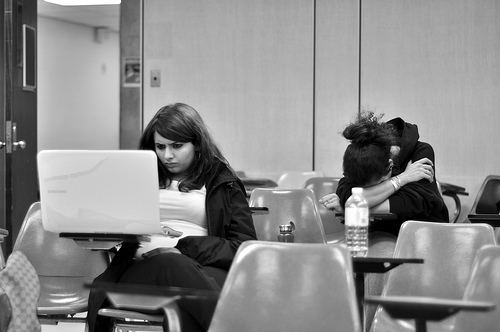}\\[0.6mm]
  {\small Original image}
\end{minipage}
\hfill
\begin{minipage}[t]{0.48\textwidth}
  \centering
  \includegraphics[width=\linewidth,height=3.2cm,keepaspectratio]{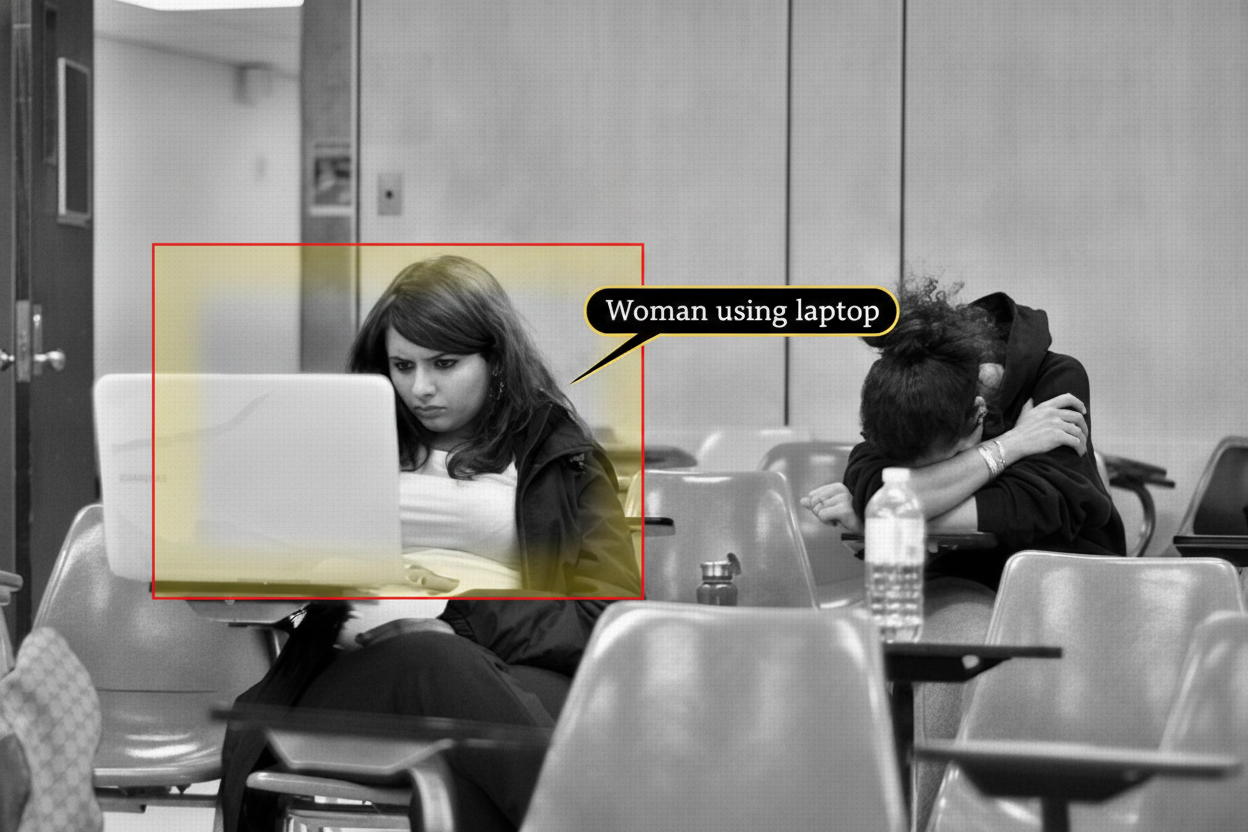}\\[0.6mm]
  {\small Rendered visual state}
\end{minipage}

\caption{\textbf{Representative successful VAoT trajectories.}
Top: highlighting the parallel branches supports equivalent-resistance
reasoning. Middle: marking the target bus supports relation-based counting.
Bottom: grounding the woman and laptop supports activity recognition.}
\label{fig:qualitative_success}
\end{figure}

Figure~\ref{fig:qualitative_success} shows cases in which the requested action
is task-relevant, faithfully rendered, and explicitly used in subsequent
reasoning. Each trajectory receives Action Relevance, Render Faithfulness, and
Feedback Uptake scores of $1$.

\subsection{Process-level Failure Cases}
\label{app:qualitative_failures}

\begin{figure}[H]
\centering
\textbf{2D structured reasoning: convex-hull diagnosis (GPT-o3)}\par\vspace{2mm}
\begin{minipage}[t]{0.48\textwidth}
  \centering
  \includegraphics[width=\linewidth,height=3.2cm,keepaspectratio]{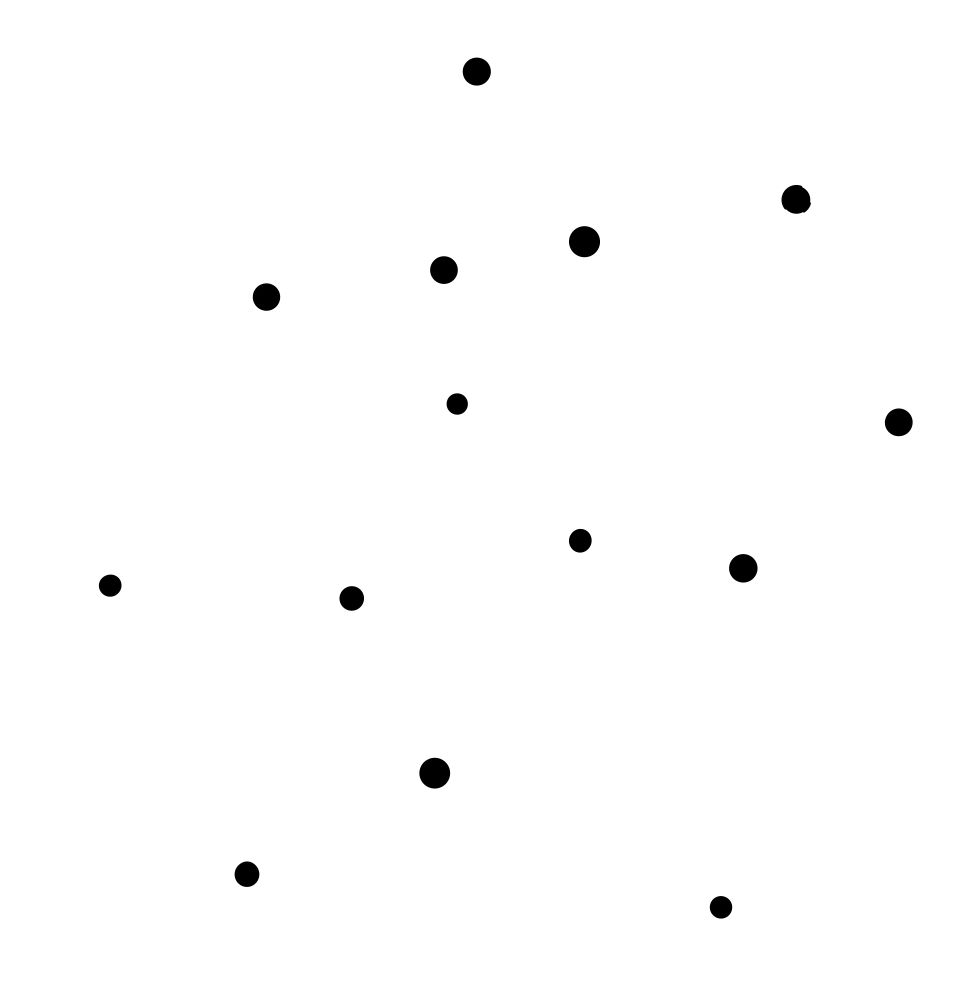}\\[0.6mm]
  {\small Original image}
\end{minipage}
\hfill
\begin{minipage}[t]{0.48\textwidth}
  \centering
  \includegraphics[width=\linewidth,height=3.2cm,keepaspectratio]{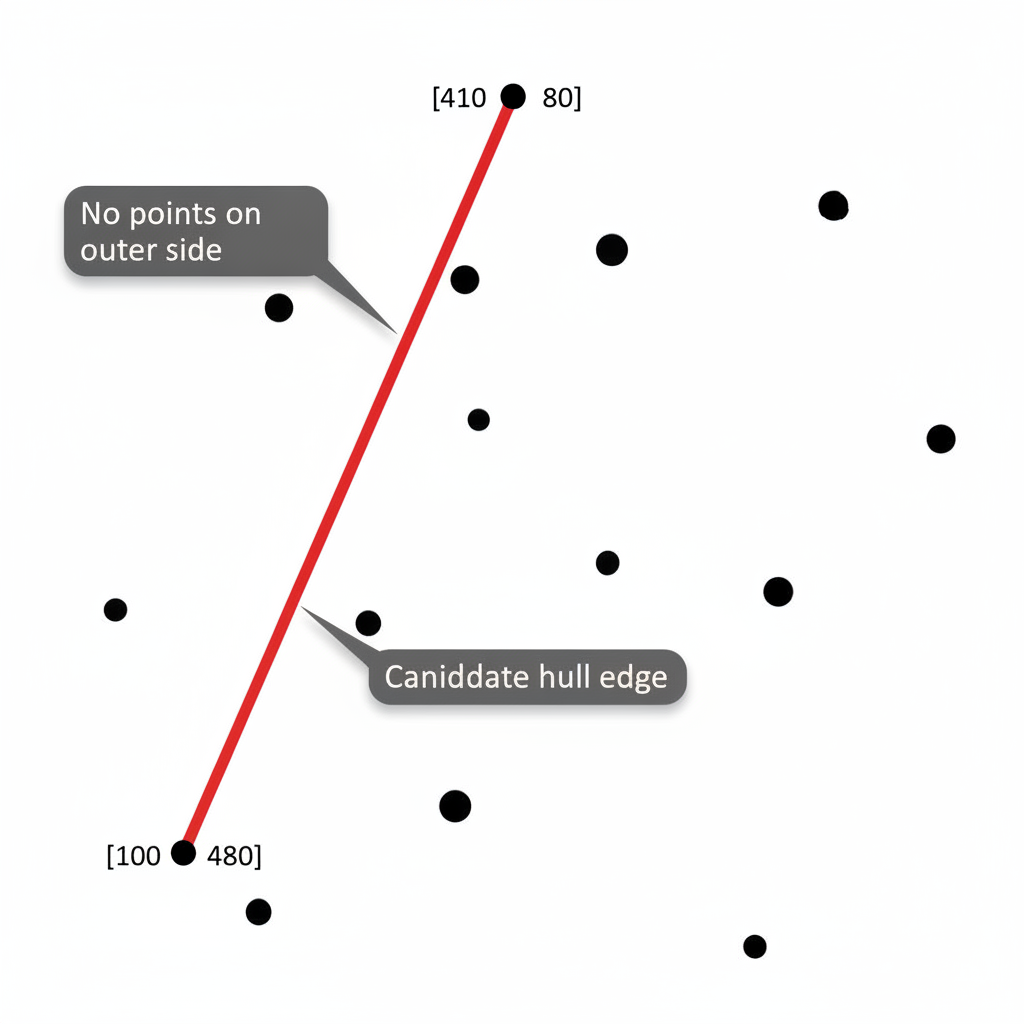}\\[0.6mm]
  {\small Ambiguous rendered state}
\end{minipage}

\vspace{4mm}
\textbf{3D scene reasoning: material identification (Gemini 3.5 Flash)}\par\vspace{2mm}
\begin{minipage}[t]{0.48\textwidth}
  \centering
  \includegraphics[width=\linewidth,height=3.2cm,keepaspectratio]{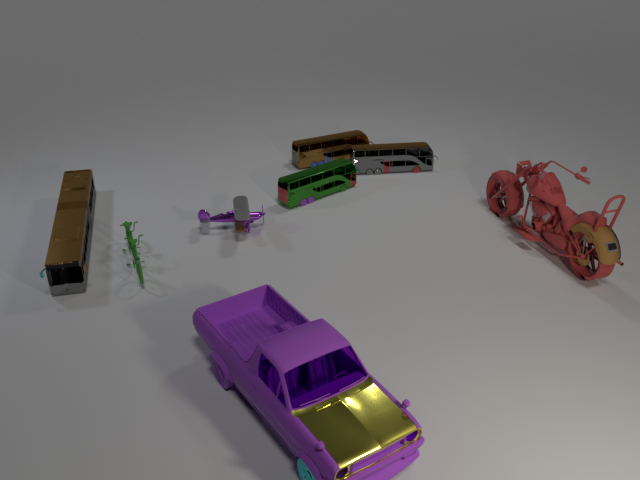}\\[0.6mm]
  {\small Original image}
\end{minipage}
\hfill
\begin{minipage}[t]{0.48\textwidth}
  \centering
  \includegraphics[width=\linewidth,height=3.2cm,keepaspectratio]{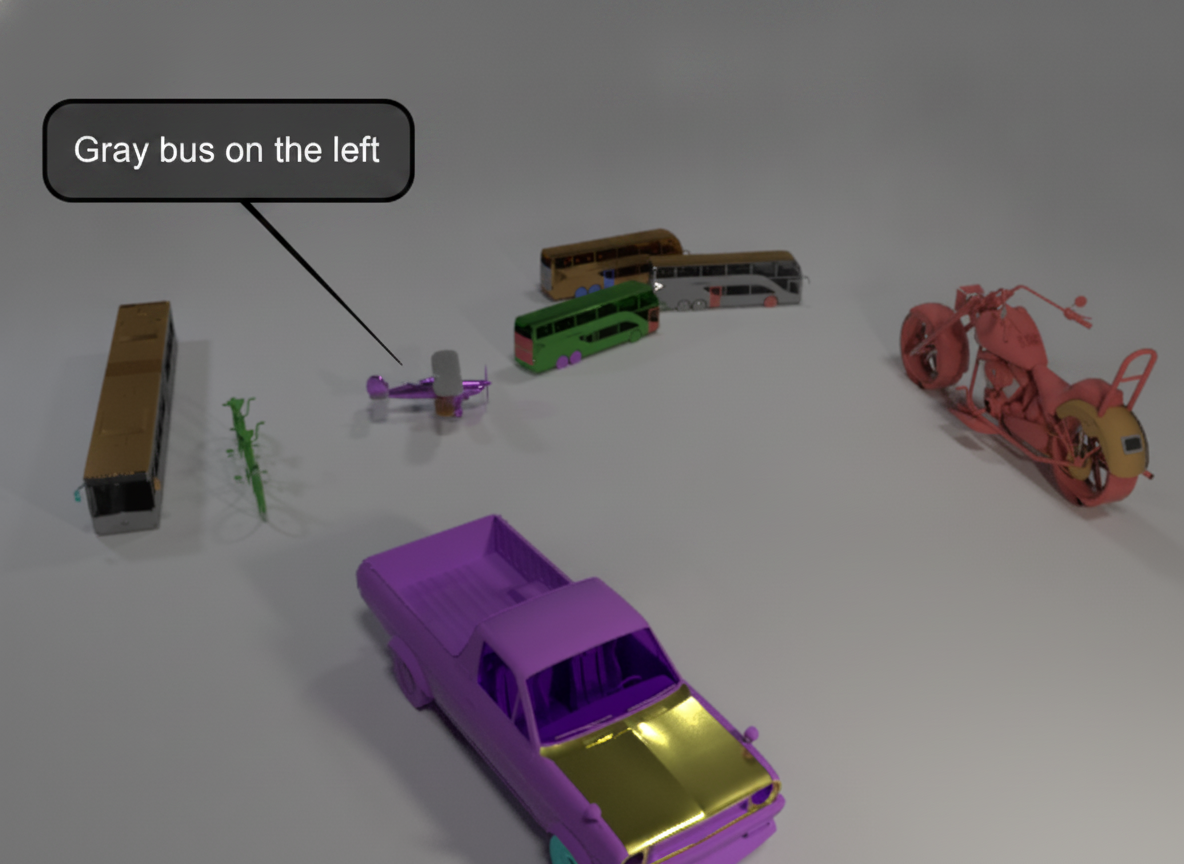}\\[0.6mm]
  {\small Misaligned rendered state}
\end{minipage}

\vspace{4mm}
\textbf{Real-world reasoning: robot manipulation (GPT-5.5)}\par\vspace{2mm}
\begin{minipage}[t]{0.48\textwidth}
  \centering
  \includegraphics[width=\linewidth,height=3.0cm,keepaspectratio]{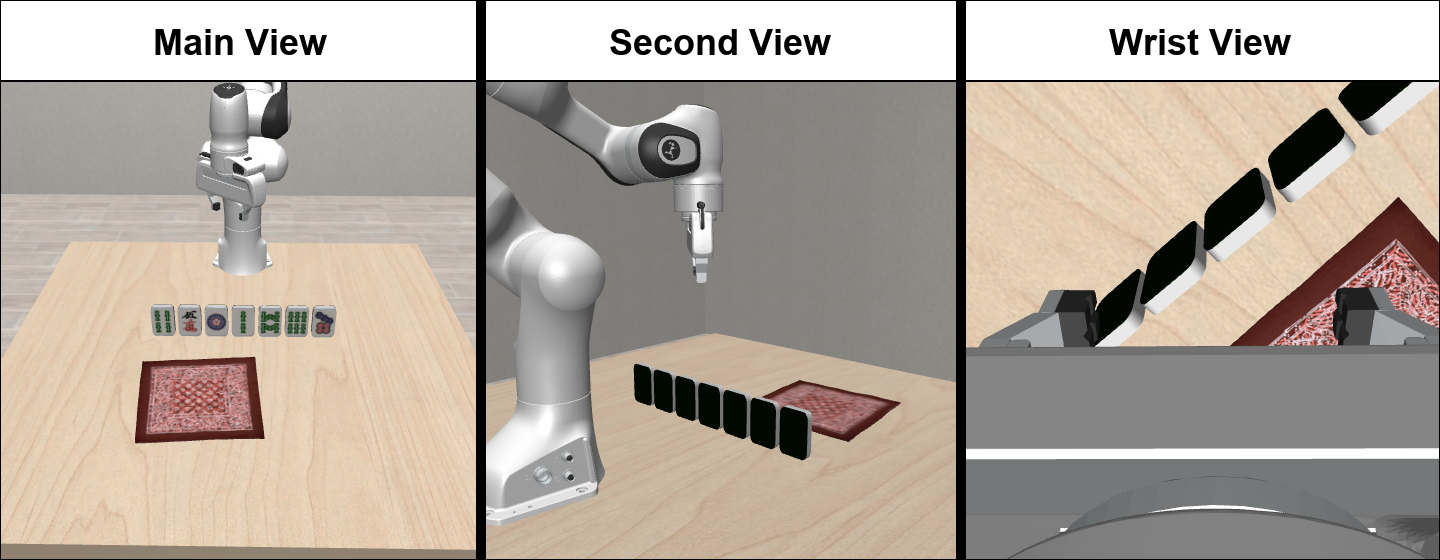}\\[0.6mm]
  {\small Original image}
\end{minipage}
\hfill
\begin{minipage}[t]{0.48\textwidth}
  \centering
  \includegraphics[width=\linewidth,height=3.0cm,keepaspectratio]{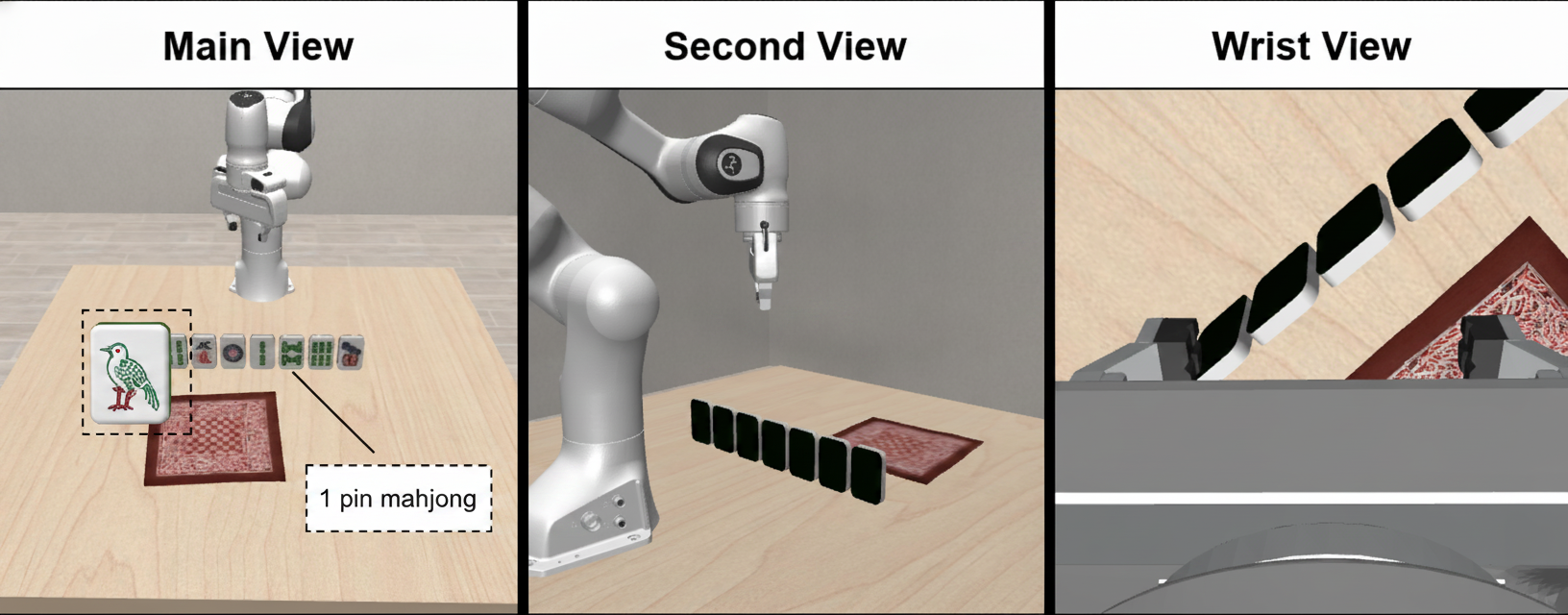}\\[0.6mm]
  {\small Partially faithful rendered state}
\end{minipage}

\caption{\textbf{Representative process-level failure cases.}
The examples separate a relevant action proposal from the quality of its
execution and subsequent use, illustrating why high Action Relevance alone
does not guarantee a correct answer.}
\label{fig:qualitative_failures}
\end{figure}

Figure~\ref{fig:qualitative_failures} illustrates distinct failure modes after
a relevant action is proposed. The 2D case is only partially rendered and
partially used $(1,0.5,0.5)$; the 3D case marks the wrong object and is not used
reliably $(1,0,0)$; and the robot-manipulation case is only partially faithful
and receives no measurable uptake $(1,0.5,0)$, contributing to an incorrect
continuous action prediction.

\subsection{Answer Changes under WrongRender}
\label{app:qualitative_wrongrender}

\begin{figure}[H]
\centering
\textbf{2D structured reasoning: prism geometry (GPT-5.5)}\par\vspace{2mm}
\begin{minipage}[t]{0.315\textwidth}
  \centering
  \includegraphics[width=\linewidth,height=2.4cm,keepaspectratio]{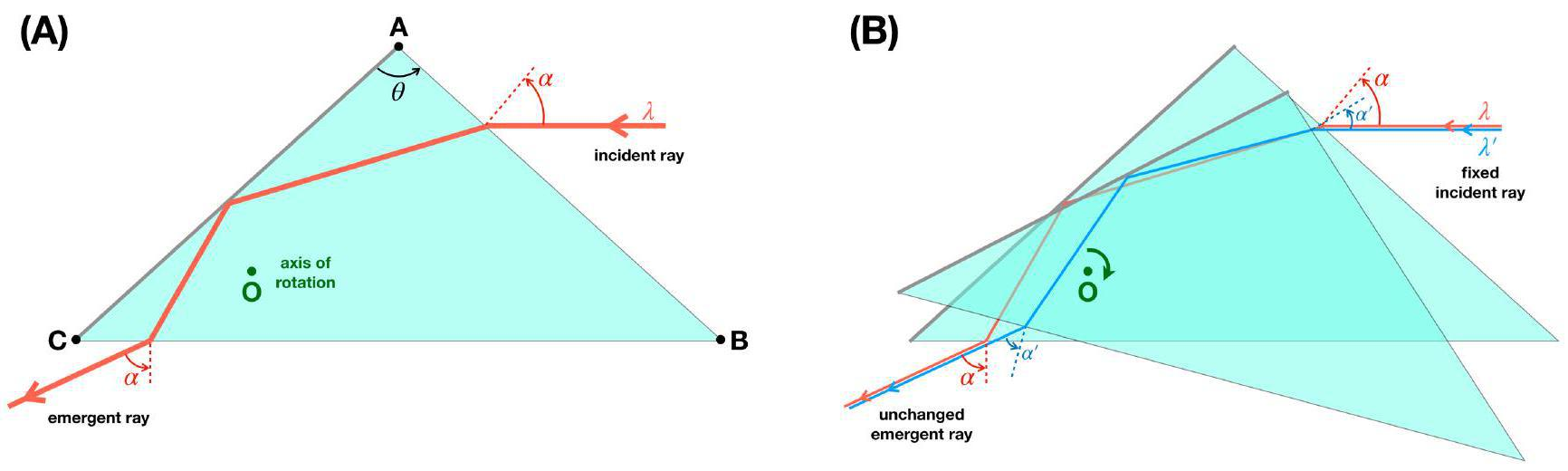}\\[0.6mm]
  {\small Original}
\end{minipage}
\hfill
\begin{minipage}[t]{0.315\textwidth}
  \centering
  \includegraphics[width=\linewidth,height=2.4cm,keepaspectratio]{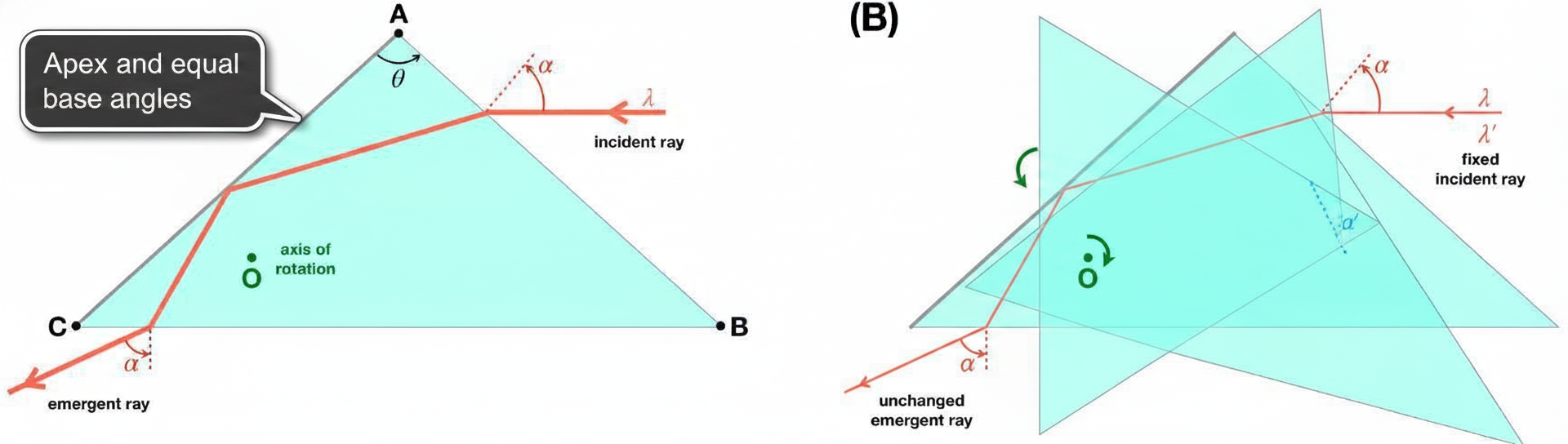}\\[0.6mm]
  {\small Faithful render}
\end{minipage}
\hfill
\begin{minipage}[t]{0.315\textwidth}
  \centering
  \includegraphics[width=\linewidth,height=2.4cm,keepaspectratio]{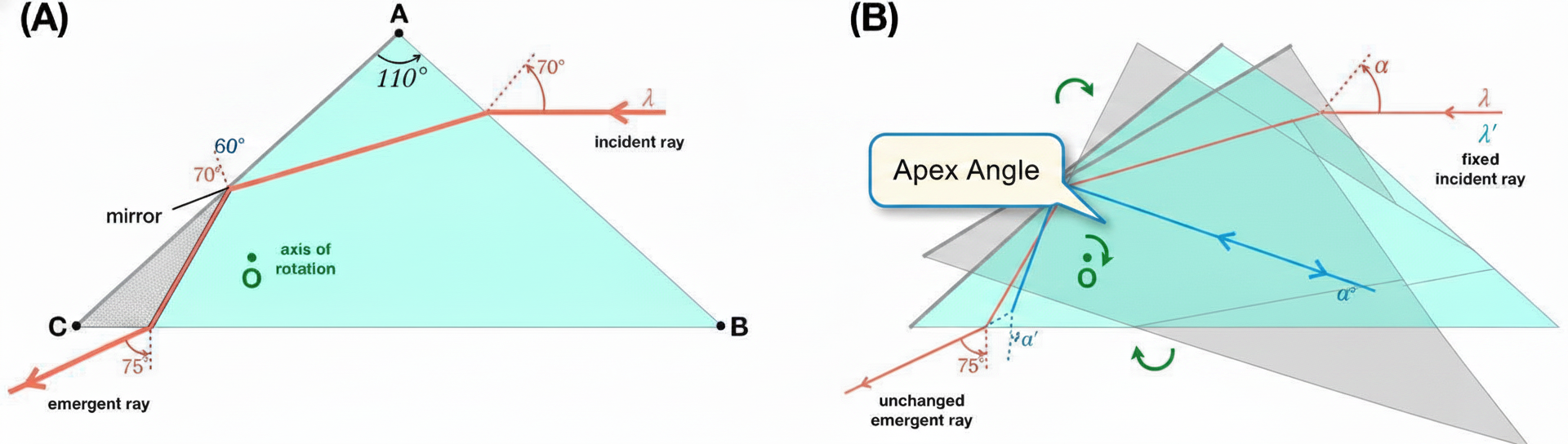}\\[0.6mm]
  {\small WrongRender}
\end{minipage}

\vspace{4mm}
\textbf{3D scene reasoning: relational object counting (GPT-o3)}\par\vspace{2mm}
\begin{minipage}[t]{0.315\textwidth}
  \centering
  \includegraphics[width=\linewidth,height=3.0cm,keepaspectratio]{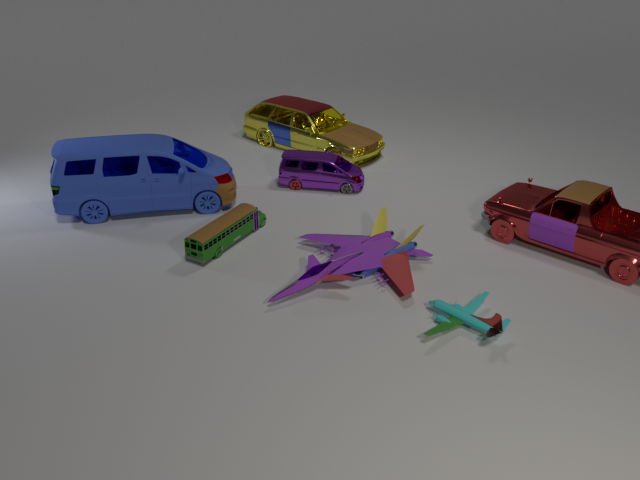}\\[0.6mm]
  {\small Original}
\end{minipage}
\hfill
\begin{minipage}[t]{0.315\textwidth}
  \centering
  \includegraphics[width=\linewidth,height=3.0cm,keepaspectratio]{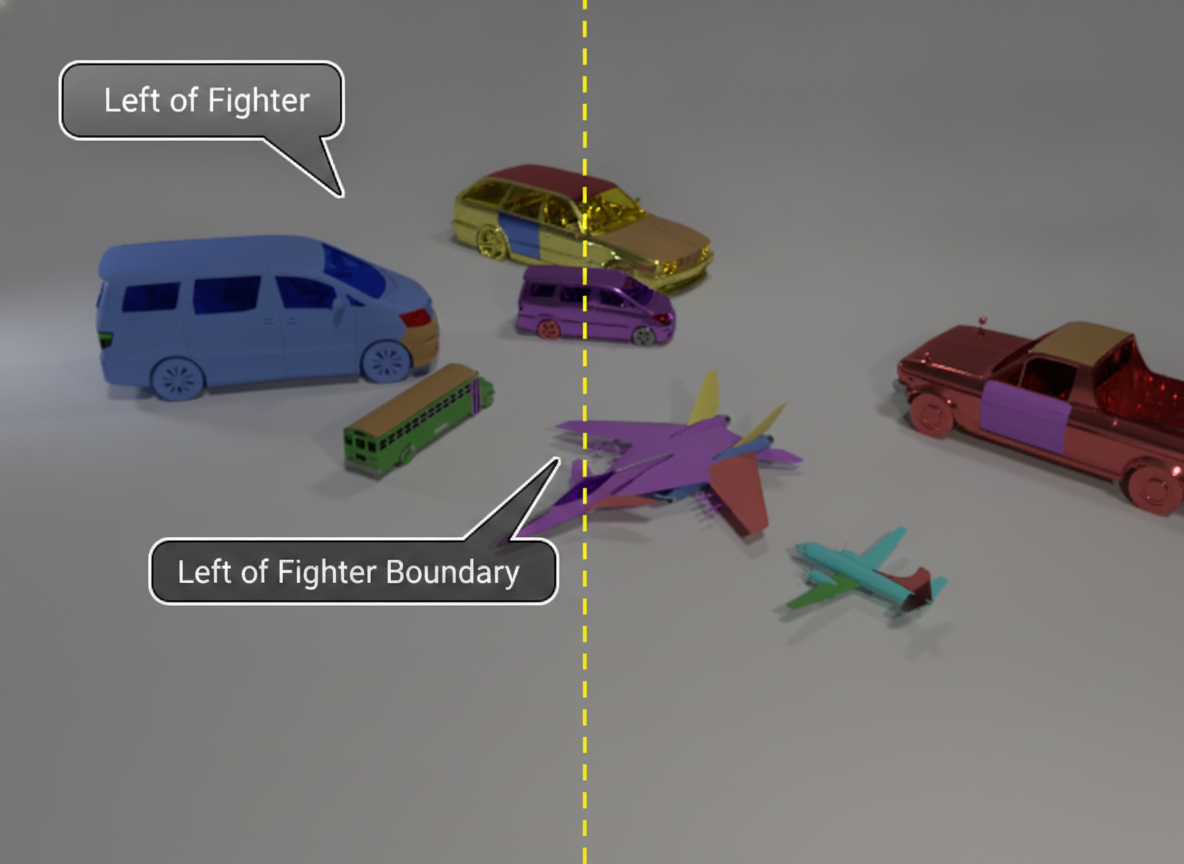}\\[0.6mm]
  {\small Faithful render}
\end{minipage}
\hfill
\begin{minipage}[t]{0.315\textwidth}
  \centering
  \includegraphics[width=\linewidth,height=3.0cm,keepaspectratio]{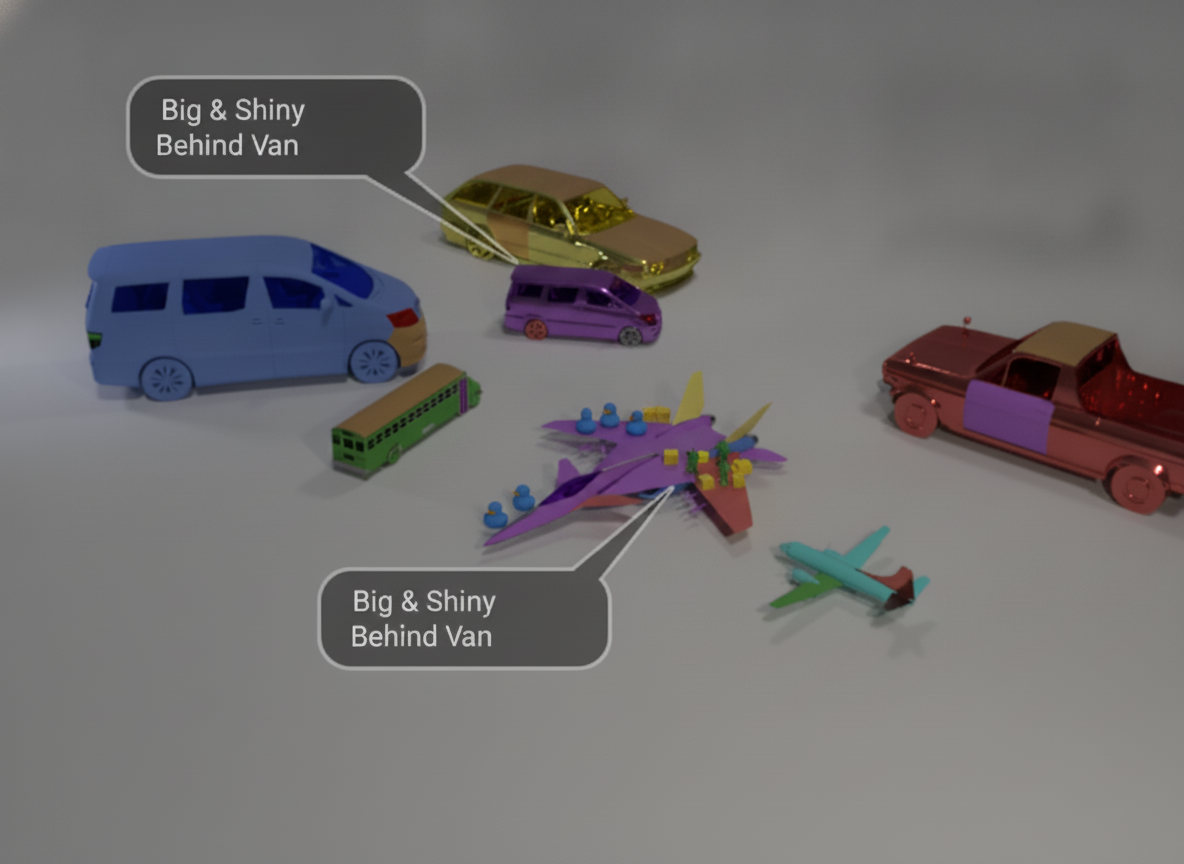}\\[0.6mm]
  {\small WrongRender}
\end{minipage}

\vspace{4mm}
\textbf{Real-world reasoning: branded-tie interpretation (Gemini 3.5 Flash)}\par\vspace{2mm}
\begin{minipage}[t]{0.315\textwidth}
  \centering
  \includegraphics[width=\linewidth,height=3.8cm,keepaspectratio]{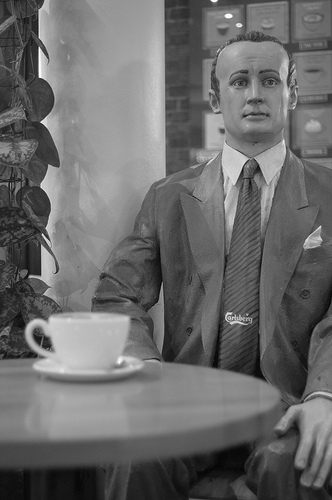}\\[0.6mm]
  {\small Original}
\end{minipage}
\hfill
\begin{minipage}[t]{0.315\textwidth}
  \centering
  \includegraphics[width=\linewidth,height=3.8cm,keepaspectratio]{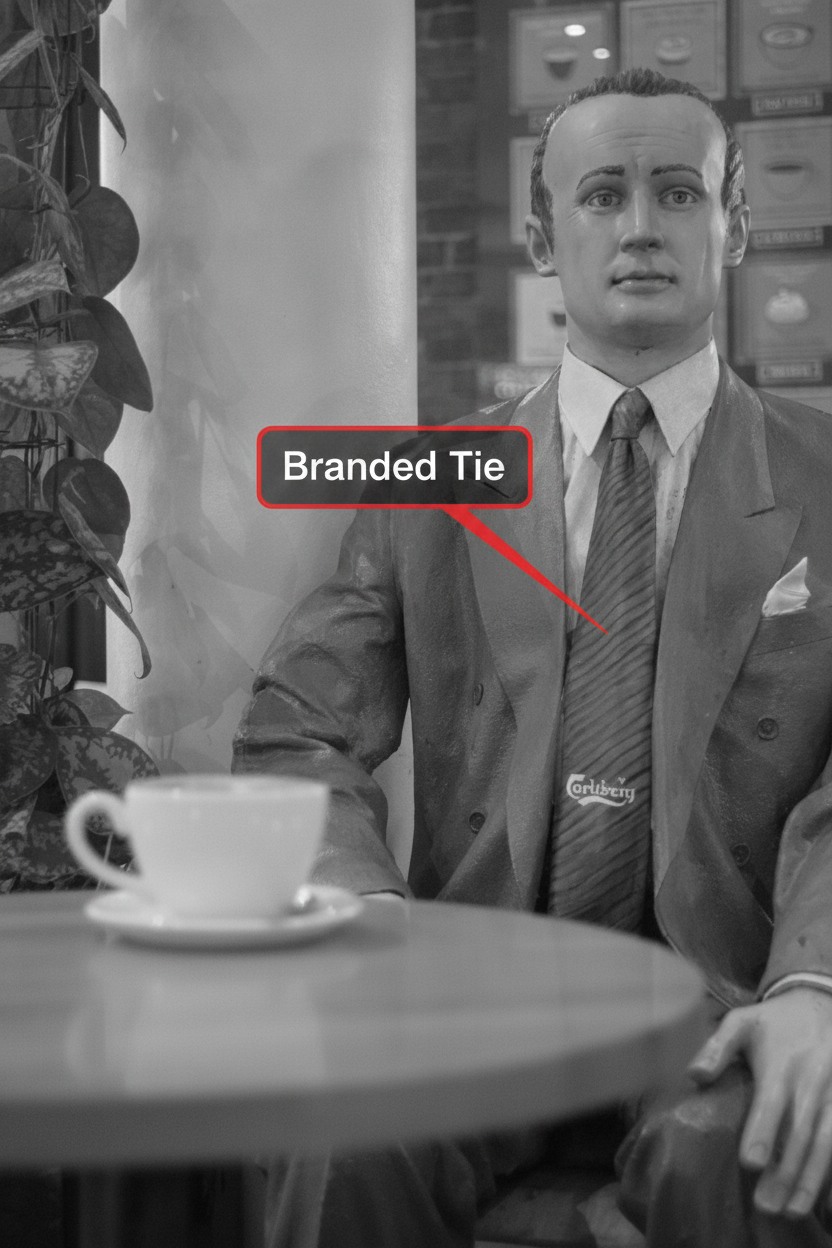}\\[0.6mm]
  {\small Faithful render}
\end{minipage}
\hfill
\begin{minipage}[t]{0.315\textwidth}
  \centering
  \includegraphics[width=\linewidth,height=3.8cm,keepaspectratio]{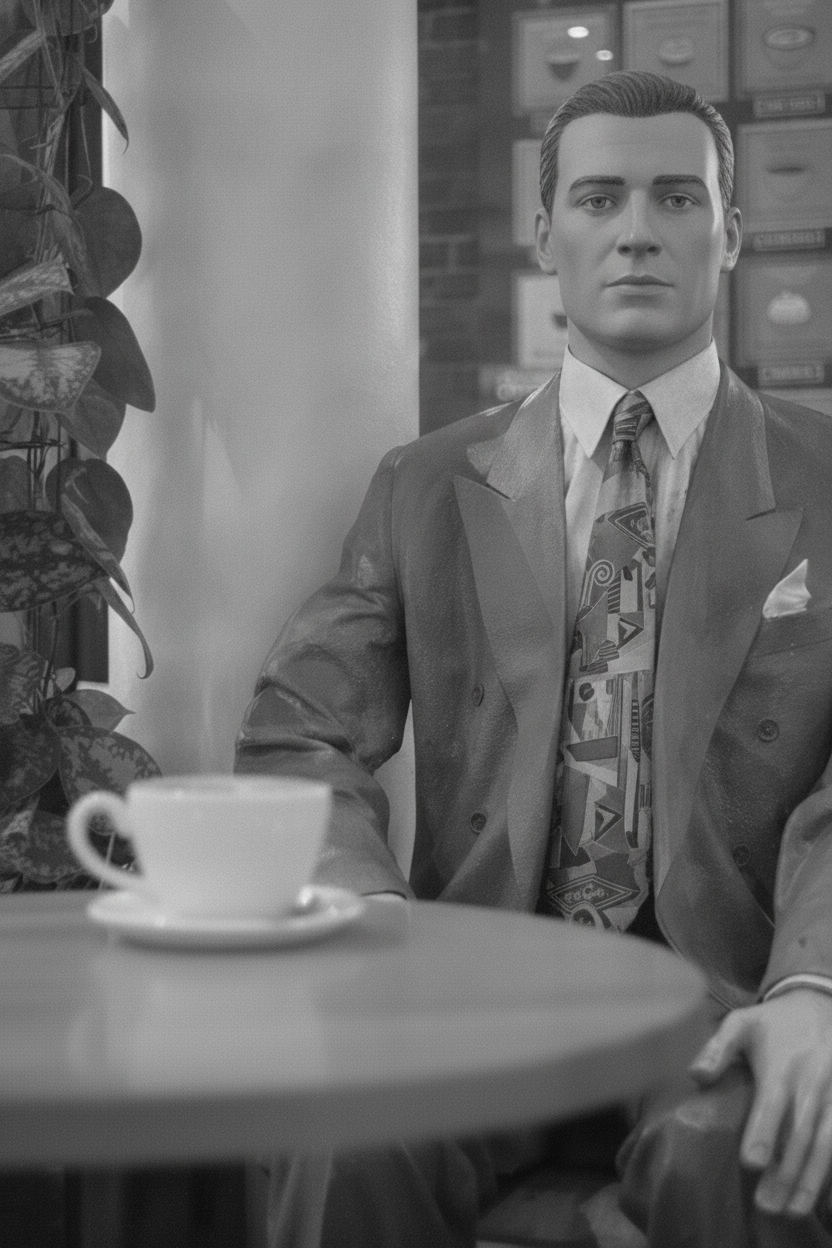}\\[0.6mm]
  {\small WrongRender}
\end{minipage}

\caption{\textbf{Representative WrongRender interventions.}
Top: the corrupted prism geometry changes the inferred refractive index.
Middle: corrupted relation markers change the object count from $2$ to $5$.
Bottom: replacing the branded tie with an unbranded pattern changes the
interpretation from a promotional display to a decorative mannequin. These
examples are qualitative illustrations; aggregate intervention quality is
assessed separately by the human quality audit in
Appendix~\ref{app:wrongrender_audit}.}
\label{fig:qualitative_wrongrender}
\end{figure}

Figure~\ref{fig:qualitative_wrongrender} compares the original image, faithful
render, and corrupted render for one case from each environment. The
interventions preserve the broad operation intent while altering task-relevant
visual content: prism geometry in 2D, target relations in 3D, and branding on a
tie in a real-world scene. In each case, the corrupted state changes the
subsequent reasoning or final answer.

\FloatBarrier

\end{document}